\pdfoutput=1

\documentclass[11pt]{article}
\usepackage{acl}
\usepackage{times}
\usepackage{latexsym}
\usepackage[utf8]{inputenc}
\usepackage[T1]{fontenc}
\usepackage{graphicx}
\usepackage{booktabs}
\usepackage{amsmath}
\usepackage[table]{xcolor}
\usepackage{placeins}
\usepackage{amssymb}
\usepackage{multirow}
\usepackage{array}
\usepackage{xcolor}
\usepackage{listings}
\usepackage[varqu,varl]{inconsolata}
\usepackage{tikz}
\usepackage{float}
\usepackage{url}
\usepackage{tabularx}
\usepackage{array}
\usepackage{booktabs}
\usepackage{tabularx}
\usepackage{array}
\usepackage{makecell}
\usepackage{booktabs}

\newcommand{\HToneN}{100}

\newcommand{\HTQHmean}{0.92}
\newcommand{\HTQOmni}{0.85}
\newcommand{\HTQKapHH}{0.96}
\newcommand{\HTQKapOH}{0.74}
\newcommand{\HTQOmniRecPos}{0.86}
\newcommand{\HTQOmniRecNeg}{0.84}

\newcommand{\HTIHmean}{0.92}
\newcommand{\HTIOmni}{0.76}
\newcommand{\HTIKapHH}{0.90}
\newcommand{\HTIKapOH}{0.55}
\newcommand{\HTKapOHLo}{0.55}
\newcommand{\HTKapOHHi}{0.76}
\newcommand{\HTIOmniRecPos}{0.88}
\newcommand{\HTIOmniRecNeg}{0.64}

\definecolor{chipposbg}{RGB}{232,241,250}\definecolor{chipposfg}{RGB}{23,88,166}
\definecolor{chipnegbg}{RGB}{253,236,226}\definecolor{chipnegfg}{RGB}{178,74,28}
\definecolor{chipneubg}{RGB}{240,240,240}\definecolor{chipneufg}{RGB}{92,92,92}
\newcommand{\chipbase}[3]{\begin{tikzpicture}[baseline=(X.base)]\node[rounded corners=2.4pt,fill=#1,text=#2,inner xsep=3.2pt,inner ysep=1.6pt,font=\scriptsize\sffamily] (X) {#3};\end{tikzpicture}}
\newcommand{\chipP}[1]{\chipbase{chipposbg}{chipposfg}{#1}}
\newcommand{\chipN}[1]{\chipbase{chipnegbg}{chipnegfg}{#1}}
\newcommand{\chipZ}[1]{\chipbase{chipneubg}{chipneufg}{#1}}

\usepackage[skins,breakable]{tcolorbox}
\newtcolorbox{promptcard}[1]{enhanced jigsaw, breakable, colback=promptbg, colframe=promptbar,
  boxrule=0.7pt, arc=2mm, left=8pt, right=8pt, top=7pt, bottom=6pt,
  title={#1}, coltitle=white, colbacktitle=promptbar,
  fonttitle=\sffamily\bfseries\small, toptitle=3pt, bottomtitle=3pt}
\newcommand{\pblock}[1]{\par\smallskip\noindent\chipbase{promptbar}{white}{#1}\par\vspace{-1pt}}
\lstdefinestyle{cardcode}{
  basicstyle=\fontsize{9.5pt}{11.3pt}\selectfont\ttfamily,
  breaklines=true, breakindent=0pt, columns=fullflexible, keepspaces=true,
  upquote=true, frame=none, aboveskip=3pt, belowskip=1pt, xleftmargin=0pt}

\newcommand{\dlgcont}{\hspace*{11pt}}

\definecolor{notegray}{RGB}{112,112,112}

\newcommand{\cnote}[1]{\par\addvspace{4pt}\noindent\hfil
  \begin{minipage}{0.86\linewidth}\centering\footnotesize\itshape\color{notegray}#1\end{minipage}\hfil\par\addvspace{2pt}}
\newcommand{\cmpU}[1]{{\small\textbf{User:}~#1}}
\newcommand{\cmpA}[1]{{\small\textbf{Assistant:}~#1}}

\definecolor{promptbar}{RGB}{42,120,214}
\definecolor{promptbg}{RGB}{247,249,251}
\lstdefinestyle{prompt}{
  basicstyle=\fontsize{8.2pt}{10pt}\selectfont\ttfamily,
  breaklines=true,
  breakindent=0pt,
  columns=fullflexible,
  keepspaces=true,
  upquote=true,
  frame=leftline,
  framerule=1.6pt,
  rulecolor=\color{promptbar},
  backgroundcolor=\color{promptbg},
  xleftmargin=9pt,
  framexleftmargin=5pt,
  xrightmargin=2pt,
  aboveskip=7pt,
  belowskip=7pt,
}

\title{Hear2Act: Benchmarking When Prosody Should Change\\What an Assistant Does}
\author{%
  \textbf{Xinyi Liu\textsuperscript{1,2}}\thanks{Work done during an internship at Amazon.} \quad
  \textbf{Hooshang Nayyeri\textsuperscript{1}} \quad
  \textbf{Dilek Hakkani-Tur\textsuperscript{1,2}} \quad
  \textbf{Emine Yilmaz\textsuperscript{1,3}} \\
  \textbf{Joo-Kyung Kim\textsuperscript{1}} \quad
  \textbf{Yifei Zhang\textsuperscript{1}} \quad
  \textbf{Charith Peris\textsuperscript{1}} \quad
  \textbf{Hari Thadakamalla\textsuperscript{1}} \\[0.45em]
  \textsuperscript{1}Amazon \quad
  \textsuperscript{2}University of Illinois Urbana-Champaign \quad
  \textsuperscript{3}University College London \\[0.15em]
  \texttt{\{liu323, dilek\}@illinois.edu} \\
  \texttt{\{hooshang, jookyk, jimmyzyf, perisc, thadakah\}@amazon.com} \\
  \texttt{eminey@amazon.co.uk}
}
\begin{document}
\setcounter{secnumdepth}{2}
\maketitle

\begin{abstract}
Prosodic cues can carry task-relevant information that words omit, but existing benchmarks rarely test whether it changes downstream decisions. We introduce \textsc{Hear2Act}, a 480-scenario protocol with unstated user needs and verifiable task success. Across matched scenarios, we keep the task and user needs fixed, varying whether an unmet need is stated in words or absent from the transcript while the user’s reaction is conveyed through prosody, yielding Explicit lexical and Prosody-mediated conditions. We vary assistant access to transcript, audio, and an indication of whether the recommendation is resolved, and evaluate two audio-capable LLMs and five text LLMs on behavior and task outcomes.

Our results show an asymmetry: prosodic information is valuable when the need is absent from the transcript, but adds little when that need is stated in words. Yet neither audio-capable LLM benefits much from raw audio: adding audio to the transcript changes task success only from 14.3\% to 15.1\%. When models infer from audio whether the recommendation remains unresolved and use that inference with the transcript, success rises to 39.6\%, close to 40.5\% with ground-truth state. This reveals a perception--action gap: raw audio barely helps, while making the information explicit in text enables much better decisions.

\end{abstract}

\section{Introduction}
\label{sec:intro}

Task-oriented dialogue (TOD) systems help users complete goal-oriented tasks through multi-turn interaction~\citep{budzianowski2018multiwoz,rastogi2020sgd,si2023spokenwoz}. In spoken TOD, the assistant's actions and final recommendation may depend on information beyond the transcript. A hesitant ``Okay, sounds good'' may indicate that an important need remains unmet, while the same words delivered brightly may signal genuine acceptance. This distinction can determine whether the assistant continues gathering information or prematurely commits to an unsuitable solution. The central evaluation question is therefore when prosodic information adds decision value beyond the transcript, and whether assistants can use it to improve information gathering and final selection.

\begin{figure*}[!t]
\centering
\includegraphics[width=0.94\textwidth]{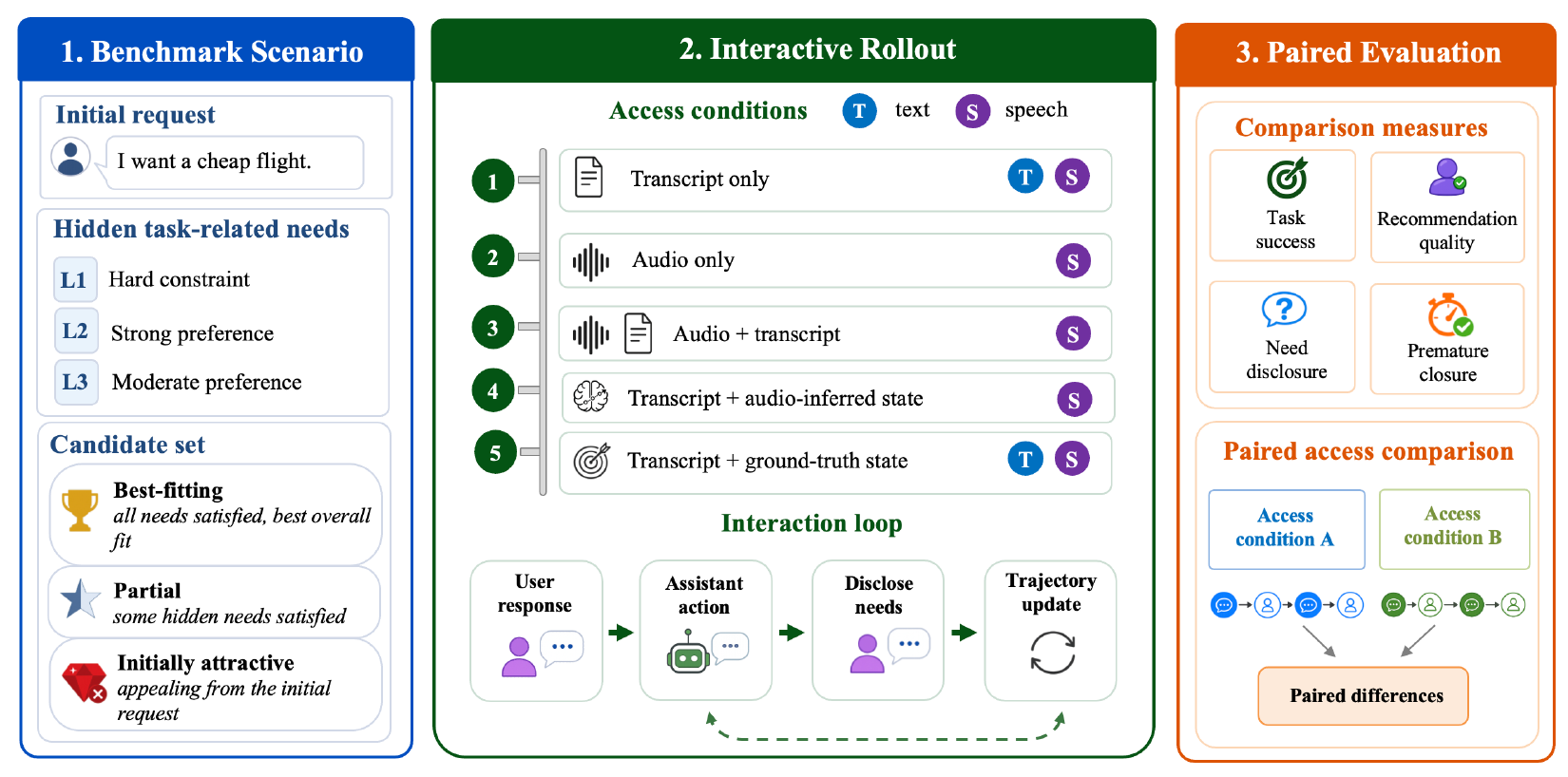}
\caption{\textbf{\textsc{Hear2Act} overview.}
(1) Each scenario combines an initial request, prioritized hidden needs, and candidates with known need-satisfaction profiles.
(2) The assistant interacts under different response and access conditions.
(3) Matched rollouts are compared on task outcomes and interaction behavior.}
\label{fig:framework}
\end{figure*}

Existing benchmarks cover complementary parts of this problem, but not the full path from prosodic evidence to task outcomes. Task-oriented dialogue benchmarks support multi-turn interaction and verifiable success, but do not isolate how prosody affects sequential decisions~\citep{budzianowski2018multiwoz,rastogi2020sgd,si2023spokenwoz}. Paralinguistic benchmarks evaluate perception, response appropriateness, or emotion-aware dialogue, but generally lack task-grounded unstated needs and objectively verifiable outcomes~\citep{ao2024sdeval,gao2024adubench,multibench2025,humdial2026}. The closest controlled settings, StyleTalk and ParaS2S, hold lexical content fixed while varying speaking style, but evaluate only the next response rather than downstream multi-turn task behavior~\citep{lin2024styletalk,paras2s2025}. To our knowledge, no existing protocol traces prosodic evidence through multi-turn task decisions to objectively verifiable outcomes.

We introduce \textsc{Hear2Act} to connect these pieces in a single matched evaluation, summarized in Figure~\ref{fig:framework}. Each of its 480 scenarios specifies an initial request, three prioritized but initially unstated needs, and candidate options with known fit to those needs. During interaction, the user responds to each recommendation and reveals unmet needs incrementally through the dialogue. Crucially, the same scenario can be realized in two ways: an unmet need is either stated explicitly, or remains absent from the transcript while the user's reaction is conveyed through prosody. We then vary the evidence available to the assistant while holding the underlying task and user needs fixed. This matched design lets us measure how prosodic information affects information gathering, stopping, and final selection, and whether it improves task success.

Applying this protocol to two audio-capable LLMs and five text LLMs yields three main findings. First, prosodic information has much greater decision value when the relevant need is absent from the transcript. Across five text LLMs, providing the correct resolved/unresolved state raises task success by 36.8 percentage points under Prosody-mediated responses, compared with only 6.3 points when the unmet need is stated explicitly. Second, current spoken models perceive more than they use. The primary spoken model recovers the intended state with 0.85 accuracy in a separate perception test, yet adding raw audio to the transcript raises average task success across the two audio-capable LLMs only from 14.3\% to 15.1\%. When models instead infer the state from audio and provide that inference in text to the downstream decision stage, success rises to 39.6\%, close to 40.5\% with the ground-truth state. Third, correct state information changes how assistants interact, not just what they ultimately choose. Across the five text LLMs, full disclosure rises from 3\% to 30\%, while premature closure falls from 69.7\% to 26.7\%. Together, these results reveal a perception--action gap: the relevant cue is recoverable from speech, but current models struggle to turn it into effective task behavior.

\section{Related Work}
\label{sec:related}

\begin{table}[!t]
\centering
\caption{\textbf{Benchmark positioning.}
S: spoken input,
C: matched control of prosodic access with fixed lexical content,
M: multi-turn task decisions,
N: task-grounded hidden user need,
and O: verifiable trajectory and outcome.
$\triangle$ marks structured user goals conveyed lexically rather than hidden needs.}
\label{tab:position}
\vspace{2pt}
\small
\setlength{\tabcolsep}{2.6pt}
\renewcommand{\arraystretch}{1.18}
\begin{tabular}{@{}>{\raggedright\arraybackslash}p{4.35cm} ccccc@{}}
\toprule
\textbf{Benchmark} & S & C & M & N & O \\
\midrule
MultiWOZ, SGD
{\scriptsize\citep{budzianowski2018multiwoz,rastogi2020sgd}}
& {\color{black!40}$\times$}
& {\color{black!40}$\times$}
& \checkmark
& $\triangle$
& \checkmark \\

SpokenWOZ
{\scriptsize\citep{si2023spokenwoz}}
& \checkmark
& {\color{black!40}$\times$}
& \checkmark
& $\triangle$
& \checkmark \\

StyleTalk, ParaS2S
{\scriptsize\citep{lin2024styletalk,paras2s2025}}
& \checkmark
& \checkmark
& {\color{black!40}$\times$}
& {\color{black!40}$\times$}
& {\color{black!40}$\times$} \\

MULTI-Bench, HumDial-EIBench
{\scriptsize\citep{multibench2025,humdial2026}}
& \checkmark
& {\color{black!40}$\times$}
& \checkmark
& {\color{black!40}$\times$}
& {\color{black!40}$\times$} \\
\midrule

\textbf{Hear2Act (ours)}
& \checkmark
& \checkmark
& \checkmark
& \checkmark
& \checkmark \\
\bottomrule
\end{tabular}
\end{table}

\begin{figure*}[!t] \centering
\includegraphics[width=0.9\textwidth]{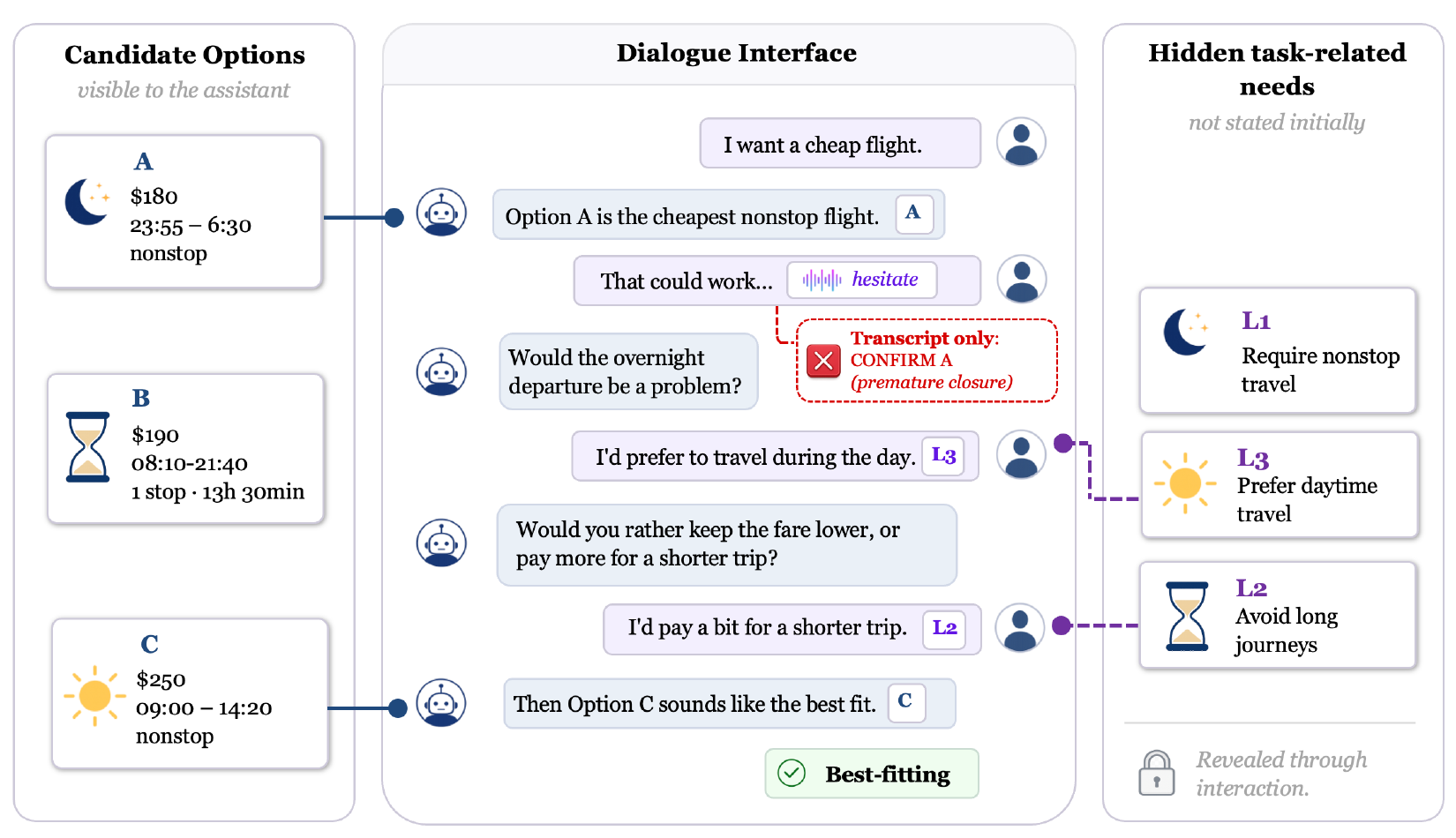} 
\caption{\textbf{Illustrative Hear2Act trajectory under Prosody-mediated responses.}
Three representative candidates are shown from the full 11-candidate set. Option A violates only the L3 preference, which remains unstated, while the user's hesitant delivery indicates that the recommendation remains unresolved. Under Explicit lexical responses, the same turn instead states the preference in words, e.g., ``That could work, though I'd prefer to travel during the day.''}
\label{fig:example} \end{figure*}

\subsection{Task-Oriented and Spoken Task Dialogue}

Task-oriented dialogue benchmarks such as MultiWOZ and the Schema-Guided Dialogue dataset provide structured multi-turn interaction with verifiable task success~\citep{budzianowski2018multiwoz,rastogi2020sgd}. Building on this setting, ATOD broadens evaluation to agentic capabilities~\citep{zhang2026atod}. For spoken interaction, SpokenWOZ brings task-oriented dialogue to speech, and RealTalk-CN studies speech--text interaction~\citep{si2023spokenwoz,realtalkcn2025}. Related work also considers user-side state: emotion-aware TOD and classical POMDP formulations model affect and latent user state~\citep{feng2022emowoz,lin2023emous,feng2024emoloop,young2013pomdp}. Together, these lines capture task success, speech, affect, and user state, but do not isolate when prosodic information adds decision value beyond the transcript or how it changes downstream task decisions.

\subsection{Paralinguistic and Spoken Evaluation}

Paralinguistic benchmarks typically focus on recognition, response appropriateness, or multi-turn affect~\citep{ao2024sdeval,yang2024airbench,huang2024dynamicsuperb,wang2025audiobench,vocalaffectbench2026,gao2024adubench,multibench2025,humdial2026}. Multi-turn affect benchmarks and emotion-conditioned simulators add interaction, but do not make information gathering, stopping, and final task selection jointly verifiable~\citep{multibench2025,humdial2026,lin2023emous,feng2024emoloop}.

Recent work further asks whether paralinguistic information carries into downstream behavior. LISTEN finds underuse of acoustic cues~\citep{chen2026listen}; PALLM links paralinguistic classification to response generation~\citep{kim2026aligning}; ParaBridge identifies a perception--behavior gap~\citep{wang2026parabridge}; and Miyazawa and Sato use recognized paralinguistic attitudes for dialogue-act prediction~\citep{miyazawa2026paralinguistic}. Among controlled settings, StyleTalk and ParaS2S are closest to ours because they hold lexical content fixed while varying speaking style, but evaluate only the next response~\citep{lin2024styletalk,paras2s2025}. \textsc{Hear2Act} combines this controlled comparison with multi-turn task decisions and objectively verifiable outcomes. Table~\ref{tab:position} summarizes these distinctions.

\section{Hear2Act Evaluation Design}
\label{sec:bench}

\textbf{Task Definition}
\label{sec:task}

\textsc{Hear2Act} evaluates whether an assistant can identify the best-fitting option through multi-turn interaction when relevant user needs are initially unstated. We evaluate both final task success and the information-gathering and stopping behavior that leads to it.

\subsection{Benchmark Scenarios}
\label{sec:scenario}

Each scenario begins with an initial request and three prioritized task-related needs that are initially unstated: a hard constraint (L1), a strong preference (L2), and a moderate preference (L3). In Figure~\ref{fig:example}, for example, the user asks for a cheap flight but also requires nonstop travel, prefers a shorter journey, and prefers daytime travel. These needs may be uncovered through subsequent interaction.

To make task success both nontrivial and objectively verifiable, we construct candidates around these three needs. The \(2^3=8\) possible satisfaction patterns are each represented by one candidate. We additionally include a safe alternative that satisfies all three needs but is less favorable on another task-relevant attribute, together with two candidates that violate an explicit requirement from the initial request, yielding 11 candidates in total. This construction creates a clear best-fitting option while ensuring that it is not necessarily identifiable from the initial request alone. An assistant that stops gathering information too early may therefore settle on a plausible but suboptimal option. Success requires uncovering the relevant needs and using them to distinguish among candidates.

We instantiate this structure across 48 service domains from the Schema-Guided Dialogue (SGD) dataset~\citep{rastogi2020sgd}, with ten scenarios per domain and 480 scenarios in total. The domains span travel, housing, healthcare, finance, and other consumer services. Within each domain, we vary urgency, budget, expertise, and other task-relevant circumstances so that similar initial requests can correspond to different underlying needs. Appendix~\ref{app:examples} provides complete rollout examples, and Appendix~\ref{app:prompts:gen} details the candidate-construction procedure.

\subsection{Interactive Rollouts}
\label{sec:interaction}

\paragraph{Rollout protocol.}
Each scenario is realized as a multi-turn dialogue in which the assistant decides whether to gather more information, recommend another option, or confirm the current one. A rollout fixes the scenario, assistant, response realization, and evidence available to the assistant, while the dialogue trajectory depends on the assistant's actions (Figure~\ref{fig:framework}, panel 2). Matched rollouts begin from the same initially attractive option. We use a compact action space to prevent assistants from bypassing information gathering by, for example, presenting multiple candidates and asking the user to choose. Text assistants may \textsc{ask}, \textsc{clarify}, \textsc{recommend}, or \textsc{confirm}; spoken assistants use \textsc{ask}, \textsc{recommend}, and \textsc{confirm}. We omit a separate \textsc{clarify} action for spoken assistants because pilot runs frequently produced repetitive clarification loops. A rollout ends when an option is accepted or the 20-turn budget is reached.

\paragraph{User simulator.}
The user simulator keeps user-side behavior stable across rollouts, so differences in outcomes reflect the assistant's decisions rather than variation in user behavior. After each assistant action, it follows a fixed rule for what information to reveal. If the assistant asks about a specific attribute, such as overnight travel, the simulator reveals the corresponding requirement or preference. If the assistant asks an open-ended question, such as whether anything else matters, it reveals only the highest-priority unmet need relevant to the current recommendation. This enforces incremental disclosure, preventing a single generic question from revealing all hidden needs. Figure~\ref{fig:example} illustrates how these needs emerge as the assistant gathers information. A language realizer then converts the resulting structured response into a natural utterance.

\begin{table}[!t]
\centering
\caption{\textbf{Hear2Act benchmark and rollout coverage.}
The 480 model-independent scenarios expand to 54,240 evaluation rollouts across models, access conditions, renderers, and interventions.}
\label{tab:stats}

\vspace{2pt}
\small
\setlength{\tabcolsep}{3pt}
\renewcommand{\arraystretch}{1.12}

\begin{tabularx}{\columnwidth}{@{}>{\raggedright\arraybackslash}X r@{}}
\toprule

\multicolumn{2}{@{}l}{\textbf{Benchmark artifact}} \\
\midrule
SGD-seeded domains                                   & 48 \\
Scenarios per domain                                 & 10 \\
\textbf{Benchmark scenarios}                         & \textbf{480} \\
Candidate options per scenario                       & 11 \\
Hidden needs per scenario                   & 3 \\
Response realizations                                & 2 \\
Base episode specifications                          & 960 \\
Assistant turn budget                                & 20 \\

\addlinespace[5pt]
\multicolumn{2}{@{}l}{\textbf{Evaluation rollouts}} \\
\midrule
Text LLM main grid                                   & 19,200 \\
Text label interventions                             & 1,440 \\
Spoken assistant with Qwen3-TTS                      & 6,720 \\
Spoken assistant with VoxCPM2                        & 6,720 \\
Qwen2-Audio, three rollouts per scenario              & 20,160 \\
\midrule
\textbf{Total evaluation rollouts}                   & \textbf{54,240} \\
\bottomrule
\end{tabularx}
\end{table}

\paragraph{Response realizations.}
We realize the same underlying user reaction in two matched ways. Under \textsc{Explicit lexical} responses, the relevant unmet need is stated in words. L1 violations trigger explicit rejection, L2 preferences are stated directly, and L3 preferences appear in qualified assent. Under \textsc{Prosody-mediated} responses, L1 violations remain explicit, but unresolved L2 and L3 needs are not stated in the transcript. Instead, vocal delivery indicates that the current recommendation remains unresolved. For example, if an option violates the user's L3 preference for daytime travel, the Explicit lexical response says, ``That could work, though I'd prefer to travel during the day.'' The matched Prosody-mediated response says only, ``That could work...'', delivered hesitantly. Thus, the latter transcript does not reveal the daytime preference itself, but the vocal delivery signals that further information gathering may be needed.

For spoken responses, the simulator assigns a target delivery such as satisfied, hesitant, concerned, or frustrated, which is realized by Qwen3-TTS~\citep{hu2026qwen3tts} or VoxCPM2~\citep{zhou2026voxcpm2}. Figure~\ref{fig:example} illustrates this matched contrast.

Response realization and assistant evidence are varied separately. Section~\ref{sec:models} defines the evidence conditions, and Appendix~\ref{app:interaction} provides the full interaction and realization rules.

\subsection{Paired Evaluation}
\label{sec:evaluation}

Each comparison pairs rollouts that share the same scenario, user needs, response realization, and task instructions, differing only in the assistant's access condition (Figure~\ref{fig:framework}, panel 3). This isolates how the available evidence affects both the final outcome and the dialogue trajectory.

We evaluate task outcomes together with information gathering, disclosure, assistant actions, and stopping behavior to distinguish effective information use from simply longer interaction. Spoken-model comparisons test how well task-relevant prosodic information is carried from audio into action. Text-model comparisons measure the effect of explicitly providing the corresponding state information.

\section{Experimental Setup}
\label{sec:setup}

The benchmark contains 480 scenarios, each evaluated under both \textsc{Explicit lexical} and \textsc{Prosody-mediated} response realizations. Across models, access conditions, renderers, and interventions, the full evaluation comprises 54,240 rollouts (Table~\ref{tab:stats}).

\subsection{Evaluated Models and Speech Realization}
\label{sec:systems}

\paragraph{Evaluated assistants.}
We evaluate two audio-capable LLMs and five text LLMs. Qwen2.5-Omni-7B~\citep{qwen25omni2025} is the primary audio-capable model, with Qwen2-Audio-7B-Instruct~\citep{chu2024qwen2audio} providing a second model. The text models are Claude Opus 4.6, Kimi K2.5, DeepSeek-V3.2, GLM-5, and Qwen3-32B~\citep{anthropic2026claudeopus46,kimiteam2026kimik25visualagentic,deepseekai2025deepseekv32,glm5team2026glm5vibecodingagentic,qwen3}. All models were accessed in July 2026 under fixed decoding settings.

\paragraph{Speech rendering.}
Qwen3-TTS~\citep{hu2026qwen3tts} is the primary speech renderer for both audio-capable models. To test robustness to the speech renderer, we additionally use VoxCPM2~\citep{zhou2026voxcpm2} with Qwen2.5-Omni. Qwen3-TTS receives a natural-language instruction specifying the target delivery, whereas VoxCPM2 uses its built-in affect control. Additional prompting and speech-rendering details are provided in Appendix~\ref{app:prompts}.

\paragraph{Speech-layer validation.}
Before downstream evaluation, we verify that the intended resolved-versus-unresolved distinction survives speech rendering. Human annotators recover the intended state with 0.92 accuracy for both Qwen3-TTS and VoxCPM2, based on 100 utterances per renderer. We then evaluate Qwen2.5-Omni-7B on the same clips to test whether the primary audio-capable model can perceive the same distinction. Full protocols and results appear in Appendix~\ref{app:tonestudy}.

\subsection{Evaluation Conditions and Interventions}
\label{sec:models}

We vary the evidence available to the assistant while keeping the interaction protocol and downstream decision instructions fixed.

\paragraph{Spoken models.}
Spoken assistants are first evaluated with direct input: transcript only ($T$), audio only ($A$), or audio plus transcript ($A+T$). We then evaluate three textualized representations of the audio. HuBERT-SUPERB-ER~\citep{hsu2021hubert,yang2021superb} and SpeechBrain wav2vec2-IEMOCAP~\citep{ravanelli2021speechbrain,baevski2020wav2vec2,busso2008iemocap} provide generic affect labels. In the audio-inferred-state condition ($T+\hat S$), the model first infers from audio whether the current recommendation is resolved or unresolved, records this state in text, and then uses it together with the transcript for downstream decision making. Each textualized representation is combined with the transcript and passed to the same downstream decision stage. Finally, $T+S$ provides the ground-truth resolved/unresolved state as a diagnostic reference.

\paragraph{Text models.}
Text LLMs are evaluated with either the transcript alone ($T$) or the transcript plus the ground-truth state ($T+S$). This comparison isolates the effect of explicit state information without requiring speech perception.

\paragraph{State-fidelity intervention.}
On a 48-scenario subset under \textsc{Prosody-mediated} responses, we compare five state-access conditions: no state label, always-\textsc{Resolved}, turn-shuffled, always-\textsc{Unresolved}, and the correct turn-level state. The three corrupted conditions preserve the presence of a state signal while varying its fidelity. This tests whether the gains depend on accurate turn-specific state information rather than simply receiving a state label.

\subsection{Evaluation Measures}
\label{sec:metrics}
\begin{table*}[!t]
\centering
\caption{\textbf{Spoken-assistant results under Prosody-mediated responses} with Qwen3-TTS. $T$, $A$, $S$, and $\hat S$ denote transcript, audio, ground-truth state, and audio-inferred state. \textbf{Average} is over the two audio-capable LLMs. Bold/underline mark the best/second-best value per column. See Table~\ref{tab:audioexp} for Explicit lexical responses and Appendix~\ref{app:audiogrid} for VoxCPM2.}
\label{tab:audio}

\vspace{2pt}
\small
\setlength{\tabcolsep}{2.2pt}
\renewcommand{\arraystretch}{1.06}

\begin{tabular}{
@{}
>{\raggedright\arraybackslash}p{0.29\textwidth}
@{\hspace{4pt}}
rrr
@{\hspace{20pt}}
rrr
@{\hspace{20pt}}
rrr
@{}
}
\toprule
& \multicolumn{3}{c}{\textbf{Qwen2.5-Omni}}
& \multicolumn{3}{c}{\textbf{Qwen2-Audio}}
& \multicolumn{3}{c}{\textbf{Average}} \\

\cmidrule(l{4pt}r{4pt}){2-4}
\cmidrule(l{4pt}r{4pt}){5-7}
\cmidrule(l{4pt}r{4pt}){8-10}

\textbf{Input / representation}
& \shortstack{\strut Success\\\%$\uparrow$}
& \shortstack{\strut Accept\\\%$\uparrow$}
& \shortstack{\strut FullFit\\\%$\uparrow$}
& \shortstack{\strut Success\\\%$\uparrow$}
& \shortstack{\strut Accept\\\%$\uparrow$}
& \shortstack{\strut FullFit\\\%$\uparrow$}
& \shortstack{\strut Success\\\%$\uparrow$}
& \shortstack{\strut Accept\\\%$\uparrow$}
& \shortstack{\strut FullFit\\\%$\uparrow$} \\
\midrule

\multicolumn{10}{@{}l}{\emph{Direct input}} \\[0.5pt]

\quad Transcript only ($T$)
& 14.8 & 22.9 & 12.1
& 13.8 & 20.2 & 12.0
& 14.3 & 21.6 & 12.1 \\

\quad Audio only ($A$)
& 16.9 & 22.9 & 13.4
& 14.8 & 22.4 & 12.1
& 15.9 & 22.7 & 12.8 \\

\quad Audio + transcript ($A+T$)
& 15.4 & 24.1 & 14.4
& 14.8 & 23.4 & 13.2
& 15.1 & 23.8 & 13.8 \\

\addlinespace[1.5pt]
\multicolumn{10}{@{}l}{\emph{Textualized prosodic representations}} \\[0.5pt]

\quad Generic affect (HuBERT)
& 35.2 & \underline{40.4} & 21.4
& 29.7 & 31.5 & 19.9
& 32.5 & 36.0 & 20.7 \\

\quad Generic affect (SpeechBrain)
& 32.1 & 38.8 & 20.6
& 25.9 & 29.3 & 18.3
& 29.0 & 34.1 & 19.5 \\

\quad Audio-inferred state ($T+\hat S$)
& \underline{42.5} & \textbf{42.1} & \underline{22.8}
& \underline{36.7} & \underline{33.7} & \underline{20.7}
& \underline{39.6} & \underline{37.9} & \underline{21.8} \\

\addlinespace[0.5pt]
\midrule

\emph{Ground-truth state ($T+S$)}
& \textbf{\textit{44.0}}
& \textbf{\textit{42.1}}
& \textbf{\textit{24.4}}
& \textbf{\textit{36.9}}
& \textbf{\textit{37.7}}
& \textbf{\textit{23.5}}
& \textbf{\textit{40.5}}
& \textbf{\textit{39.9}}
& \textbf{\textit{24.0}} \\

\bottomrule
\end{tabular}
\end{table*}
\begin{table*}[!t]
\centering
\caption{\textbf{Spoken-assistant use of state information under Explicit lexical responses} with Qwen3-TTS (Qwen2.5-Omni $n{=}480$, Qwen2-Audio $n{=}1{,}440$ per condition). Notation follows Table~\ref{tab:audio}. Differences across conditions are smaller because the unmet need is lexicalized in the transcript. Bold/underline mark column-wise highest/next-highest values.}
\label{tab:audioexp}

\vspace{2pt}
\small
\setlength{\tabcolsep}{2.2pt}
\renewcommand{\arraystretch}{1.06}

\begin{tabular}{
@{}
>{\raggedright\arraybackslash}p{0.29\textwidth}
@{\hspace{4pt}}
rrr
@{\hspace{20pt}}
rrr
@{\hspace{20pt}}
rrr
@{}
}
\toprule
& \multicolumn{3}{c}{\textbf{Qwen2.5-Omni}}
& \multicolumn{3}{c}{\textbf{Qwen2-Audio}}
& \multicolumn{3}{c}{\textbf{Average}} \\

\cmidrule(l{4pt}r{4pt}){2-4}
\cmidrule(l{4pt}r{4pt}){5-7}
\cmidrule(l{4pt}r{4pt}){8-10}

\textbf{Input / representation}
& \shortstack{\strut Success\\\%$\uparrow$}
& \shortstack{\strut Accept\\\%$\uparrow$}
& \shortstack{\strut FullFit\\\%$\uparrow$}
& \shortstack{\strut Success\\\%$\uparrow$}
& \shortstack{\strut Accept\\\%$\uparrow$}
& \shortstack{\strut FullFit\\\%$\uparrow$}
& \shortstack{\strut Success\\\%$\uparrow$}
& \shortstack{\strut Accept\\\%$\uparrow$}
& \shortstack{\strut FullFit\\\%$\uparrow$} \\
\midrule

\multicolumn{10}{@{}l}{\emph{Direct input}} \\[0.5pt]

\quad Transcript only ($T$)
& \textbf{55.8} & 45.6 & \textbf{27.0}
& 48.7 & \underline{42.3} & \textbf{26.4}
& \textbf{52.3} & \underline{44.0} & \textbf{26.7} \\

\quad Audio only ($A$)
& \underline{55.2} & 45.9 & 26.5
& 48.2 & 41.5 & 26.0
& \underline{51.7} & 43.7 & 26.3 \\

\quad Audio + transcript ($A+T$)
& 51.5 & 45.6 & \underline{26.6}
& 48.0 & 41.3 & 25.9
& 49.8 & 43.5 & 26.3 \\

\addlinespace[1.5pt]
\multicolumn{10}{@{}l}{\emph{Textualized prosodic representations}} \\[0.5pt]

\quad Generic affect (HuBERT)
& 52.5 & \underline{46.3} & 25.4
& 49.0 & 40.8 & 24.5
& 50.8 & 43.6 & 25.0 \\

\quad Generic affect (SpeechBrain)
& 52.5 & \underline{46.3} & 25.4
& 45.4 & 40.1 & 23.8
& 49.0 & 43.2 & 24.6 \\

\quad Audio-inferred state ($T+\hat S$)
& 51.9 & 45.9 & 25.3
& \underline{50.3} & 41.6 & 25.2
& 51.1 & 43.8 & 25.3 \\

\addlinespace[0.5pt]
\midrule

\emph{Ground-truth state ($T+S$)}
& \textit{49.2}
& \textbf{\textit{46.7}}
& \underline{\textit{26.6}}
& \textbf{\textit{50.8}}
& \textbf{\textit{42.5}}
& \underline{\textit{26.2}}
& \textit{50.0}
& \textbf{\textit{44.6}}
& \underline{\textit{26.4}} \\

\bottomrule
\end{tabular}
\end{table*}

We evaluate both the final task outcome and the interaction that leads to it.

\paragraph{Final outcome and recommendation quality.}
Our primary metric is \textbf{task success rate (Success\%)}, the proportion of rollouts that end with the best-fitting option. If the 20-turn limit is reached, we score the assistant's final recommendation. We separately evaluate recommendation quality at every recommendation turn. \textbf{Accept\%} is the percentage of recommendation turns in which the recommended option satisfies all stated requirements and the two higher-priority hidden needs (L1 and L2), allowing L3 to remain unmet. \textbf{FullFit\%} is the percentage of recommendation turns in which the recommended option satisfies all stated requirements and all three hidden needs. Success\% is measured per rollout, whereas Accept\% and FullFit\% are measured per recommendation turn.

\paragraph{Interaction behavior.}
For the text LLMs, we compare matched $T$ and $T+S$ rollouts to examine how state access changes information gathering and stopping behavior. We report two main interaction measures: \textbf{full disclosure rate}, the percentage of rollouts in which all three hidden needs are revealed, and \textbf{premature closure rate}, the percentage of rollouts that stop on a suboptimal option. Appendix~\ref{app:textbehavior} provides finer-grained analyses of \textsc{recommend}, \textsc{ask}, and \textsc{clarify} behavior and the number of hidden needs revealed.

All comparisons, including the 48-scenario intervention, use paired scenario-level bootstrap analysis with 2{,}000 resamples and 95\% confidence intervals. Models receive equal weight in pooled analyses, with key intervals reported in Appendix~\ref{app:ci}.

\section{Results}
\label{sec:results}

The results address three questions. First, when does prosodic information add decision value beyond the transcript? Second, when it does, can spoken assistants use it? Third, how does correct state information change the dialogue? All analyses use paired scenario-level comparisons. Paired 95\% confidence intervals for key diagnostic contrasts appear in Table~\ref{tab:ci}.

\subsection{When Does Prosodic Information Add Decision Value?}
\label{sec:rq1}

We first ask when prosodic information changes decisions beyond what is already available in the transcript. The results show that prosodic information has substantial decision value when lexical evidence is insufficient, but the evaluated audio-capable LLMs make limited use of this value directly from audio. Under Prosody-mediated responses, adding audio to the transcript changes the average success rate only from $14.3\%$ to $15.1\%$, whereas the two-stage audio-inferred-state condition raises it to $39.6\%$ (Table~\ref{tab:audio}). The gain is much smaller under Explicit lexical responses, where the unmet need is already verbalized: success rates across all seven conditions span only $6.6$ points for Qwen2.5-Omni and $5.4$ points for Qwen2-Audio (Table~\ref{tab:audioexp}).

This asymmetry is not specific to spoken models. When we isolate the effect of explicitly providing the corresponding state information to five text LLMs, the gains are much larger under Prosody-mediated responses. Ground-truth state access raises the mean success rate across the five models by $36.8$ percentage points (Table~\ref{tab:text}), compared with $6.3$ points under Explicit lexical responses. The remaining gain under Explicit lexical responses is concentrated in L3 turns expressed as qualified assent. In these cases, the preference is stated in words, but the utterance can still appear locally accepting. The state tag clarifies that the current recommendation remains unresolved. After such turns, immediate confirmation drops from $17.2\%$ with $T$ to $1.7\%$ with $T+S$. Accept\% and FullFit\% show the same asymmetry, changing by $+7.5$ and $+4.1$ points under Prosody-mediated responses but by only $-1.7$ and $+0.3$ points under Explicit lexical responses.

Together, these results show that prosodic information has substantial decision value when lexical evidence is insufficient, with much smaller additional value when the relevant need is already expressed in the transcript.
\begin{table*}[!t]
\centering
\caption{\textbf{Effect of ground-truth state access on text LLMs.}
Results use 480 matched scenarios with two runs each ($n{=}960$ per cell).
$T$ is transcript only; $T{+}S$ adds turn-level ground-truth state tags.
Bold marks the larger value in each pair; action composition appears in Figure~\ref{fig:actioncomp}.}
\label{tab:text}

\vspace{2pt}
\small
\setlength{\tabcolsep}{2.8pt}
\renewcommand{\arraystretch}{1.16}

\resizebox{0.8\textwidth}{!}{%
\begin{tabular}{
@{}l
rr @{\hspace{6pt}}
rr @{\hspace{6pt}}
rr
@{\hspace{16pt}}
rr @{\hspace{6pt}}
rr @{\hspace{6pt}}
rr@{}
}

\toprule
& \multicolumn{6}{c}{\textbf{Prosody-mediated responses}}
& \multicolumn{6}{c}{\textbf{Explicit lexical responses}} \\
\cmidrule(lr){2-7}\cmidrule(lr){8-13}

& \multicolumn{2}{c}{Success\%$\uparrow$}
& \multicolumn{2}{c}{Accept\%$\uparrow$}
& \multicolumn{2}{c}{FullFit\%$\uparrow$}
& \multicolumn{2}{c}{Success\%$\uparrow$}
& \multicolumn{2}{c}{Accept\%$\uparrow$}
& \multicolumn{2}{c}{FullFit\%$\uparrow$} \\
\cmidrule(lr){2-3}\cmidrule(lr){4-5}\cmidrule(lr){6-7}
\cmidrule(lr){8-9}\cmidrule(lr){10-11}\cmidrule(lr){12-13}

\textbf{Model}
& $T$ & $T{+}S$
& $T$ & $T{+}S$
& $T$ & $T{+}S$
& $T$ & $T{+}S$
& $T$ & $T{+}S$
& $T$ & $T{+}S$ \\
\midrule

Claude Opus 4.6
& 50.3 & \textbf{79.8}
& 37.0 & \textbf{41.1}
& 24.9 & \textbf{26.3}
& 75.5 & \textbf{82.2}
& \textbf{44.6} & 42.9
& 26.0 & \textbf{26.4} \\

Kimi K2.5
& 27.8 & \textbf{69.2}
& 26.5 & \textbf{35.7}
& 14.4 & \textbf{19.6}
& 69.4 & \textbf{74.8}
& \textbf{35.8} & 35.2
& 17.7 & 17.7 \\

GLM-5
& 22.4 & \textbf{60.2}
& 25.0 & \textbf{32.5}
& 13.1 & \textbf{15.8}
& 67.6 & \textbf{73.8}
& \textbf{35.4} & 34.8
& 15.6 & \textbf{16.1} \\

Qwen3-32B
& 21.0 & \textbf{53.1}
& 29.7 & \textbf{33.6}
& 10.4 & \textbf{13.7}
& 64.3 & \textbf{64.8}
& \textbf{42.1} & 36.5
& \textbf{12.9} & 12.5 \\

DeepSeek-V3.2
& 13.2 & \textbf{56.2}
& 15.9 & \textbf{28.5}
& 6.5 & \textbf{14.5}
& 55.1 & \textbf{67.8}
& 31.3 & \textbf{31.7}
& 14.4 & \textbf{15.3} \\

\midrule
\textbf{\textit{Five-model mean}}
& \textit{26.9} & \textbf{\textit{63.7}}
& \textit{26.8} & \textbf{\textit{34.3}}
& \textit{13.9} & \textbf{\textit{18.0}}
& \textit{66.4} & \textbf{\textit{72.7}}
& \textbf{\textit{37.8}} & \textit{36.2}
& \textit{17.3} & \textbf{\textit{17.6}} \\

\bottomrule
\end{tabular}%
}
\end{table*}
\subsection{When Prosody Matters, Can Spoken Assistants Use It?}
\label{sec:rq3}

The Prosody-mediated condition is therefore the setting in which prosodic information adds substantial value beyond the transcript. We next ask where this value is lost between perception and action. Qwen2.5-Omni shows a clear perception--action gap: on the audited perception task, it recovers the intended state with \HTQOmni{} accuracy on Qwen3-TTS and \HTIOmni{} on VoxCPM2 (Table~\ref{tab:omni_reader}), showing that the cue is recoverable from audio. Yet success rates remain similar with Transcript only ($14.8\%$), Audio only ($16.9\%$), and Audio $+$ transcript ($15.4\%$; Table~\ref{tab:audio}). Thus, recovering the cue does not by itself translate into better task decisions.

A two-stage inference-and-representation condition largely closes this gap. Qwen2.5-Omni first infers the state from audio, expresses it in text, and then uses it with the transcript for next-action selection. The success rate rises from $15.4\%$ to $42.5\%$, close to $44.0\%$ with the ground-truth state. This condition recovers most of the available state-access benefit, although it does not isolate the contributions of task-aligned inference and explicit representation.

\begin{figure}[!t]
\centering
\includegraphics[width=\columnwidth]{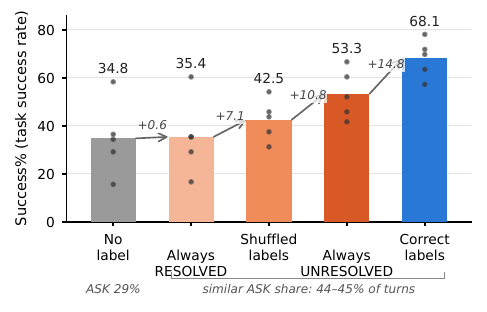}
\caption{\textbf{Outcome tracks label fidelity, not asking frequency.}
Task success rate (Success\%) on 48 Prosody-mediated intervention scenarios, pooled over five text LLMs; \textsc{ask} shares remain $44$--$45\%$.}
\label{fig:failure}
\end{figure}

We next ask whether generic speech-emotion representations recover decision value from audio. With Qwen2.5-Omni fixed as the downstream decision model, textualized HuBERT and SpeechBrain predictions raise the success rate from $15.4\%$ to $35.2\%$ and $32.1\%$, while the audio-inferred state raises it to $42.5\%$ (Table~\ref{tab:audio}). The same pattern holds for Qwen2-Audio and VoxCPM2 (Appendix~\ref{app:audiogrid}). Thus, generic affect recovers some decision value, while task-relevant state inference yields larger gains in our setting.

\subsection{How Does Useful State Information Change the Dialogue?}
\label{sec:rq2}

Correct state information shifts the dialogue from early recommendation toward information gathering. Under Prosody-mediated responses, the average number of \textsc{ask} and \textsc{clarify} actions across the five text LLMs rises from $1.8$ to $4.3$ per dialogue, while full disclosure of all three hidden needs rises from $3\%$ to $30\%$. These changes are much smaller under Explicit lexical responses. Additional behavior results appear in Appendix~\ref{app:textbehavior}.

The gain is not simply due to asking more. Incorrect state tags produce similar \textsc{ask} shares ($44$--$45\%$) but much lower success rates than the correct state ($35.4$--$53.3\%$ versus $68.1\%$; Figure~\ref{fig:failure}). The always-unresolved condition still improves over no state ($53.3\%$ versus $34.8\%$), suggesting that a generic unresolved signal discourages premature closure, while accurate turn-specific states provide additional value. Consistently, state access reduces premature closure on a suboptimal option from $69.7\%$ to $26.7\%$.

Overall, state signals encourage continued information gathering, while accurate state information further helps determine when to continue searching and when to stop.

\section{Conclusion}
\label{sec:conclusion}

We introduced \textsc{Hear2Act}, a controlled protocol for measuring when prosody changes multi-turn task decisions. Across two evaluated audio-capable LLMs, prosody adds value when lexical evidence is insufficient, yet raw audio provides little direct benefit. A two-stage inference-and-representation setup recovers most of the ground-truth-state benefit; state information also improves information gathering and stopping. Overall, the evaluated models recover useful information from speech but use it more effectively with task-aligned inference and explicit representation.

\section*{Limitations}
\label{sec:limitations}

\textsc{Hear2Act} isolates one decision-relevant function of prosody: whether and how strongly a recommendation remains unresolved. Its shared three-layer hidden-need structure and standardized opening enable matched, objectively scored comparisons, but do not cover broader preference structures, other functions of prosody, or unconstrained first-turn recommendation.

User utterances and speech are synthesized from structured states. Human validation shows that the intended state contrast remains perceptible across two renderers, but synthetic speech cannot capture the full variability of natural speech and interaction. Ground-truth-state conditions are diagnostic rather than deployment settings. The inferred-state condition adds both an explicit representation and a directed inference step, so it establishes the sufficiency of the combined intervention without isolating either component. Without a matched two-stage transcript-only inference condition, we also cannot determine how much of the gain is uniquely attributable to acoustic evidence.

\section*{Ethical Considerations}
\label{sec:ethics}

\textsc{Hear2Act} uses constructed scenarios and personas, with dialogue and speech generated from structured benchmark states. It contains no real user conversations, identities, recordings, or personal data. Personas encode only task-relevant context and are not intended to represent demographic groups or support claims about real individuals.

The benchmark evaluates a narrow communicative signal: whether a recommendation remains unresolved. Such evidence should support clarification rather than be used to infer sensitive attributes, override explicit user statements, or make high-stakes decisions without confirmation. The speech-perception validation involved graduate-student annotators who judged generated speech clips and were compensated at \$25 per hour. No personal or sensitive information was collected. Further details appear in Appendix~\ref{app:tonestudy}; benchmark and computational details appear in Appendix~\ref{app:responsible}.

\bibliography{references}

@article{busso2008iemocap,
  title={{IEMOCAP}: Interactive emotional dyadic motion capture database},
  author={Busso, Carlos and Bulut, Murtaza and Lee, Chi-Chun and Kazemzadeh, Abe and Mower, Emily and Kim, Samuel and Chang, Jeannette N and Lee, Sungbok and Narayanan, Shrikanth S},
  journal={Language Resources and Evaluation},
  volume={42},
  number={4},
  pages={335--359},
  year={2008},
  publisher={Springer}
}

@misc{anthropic2026claudeopus46,
  author = {{Anthropic}},
  title = {Claude Opus 4.6 System Card},
  year = {2026},
  url = {https://www.anthropic.com/system-cards}
}

@misc{kimiteam2026kimik25visualagentic,
  author = {{Kimi Team} and others},
  title = {Kimi K2.5: Visual Agentic Intelligence},
  year = {2026},
  eprint = {2602.02276},
  archivePrefix = {arXiv},
  primaryClass = {cs.CL},
  url = {https://arxiv.org/abs/2602.02276}
}

@misc{deepseekai2025deepseekv32,
  author = {{DeepSeek-AI}},
  title = {DeepSeek-V3.2: Pushing the Frontier of Open Large Language Models},
  year = {2025},
  eprint = {2512.02556},
  archivePrefix = {arXiv},
  primaryClass = {cs.CL},
  url = {https://arxiv.org/abs/2512.02556}
}

@misc{glm5team2026glm5vibecodingagentic,
  author = {{GLM-5 Team} and others},
  title = {{GLM}-5: from Vibe Coding to Agentic Engineering},
  year = {2026},
  eprint = {2602.15763},
  archivePrefix = {arXiv},
  primaryClass = {cs.LG},
  url = {https://arxiv.org/abs/2602.15763}
}

@article{qwen3,
  author = {An Yang and others},
  title = {Qwen3 Technical Report},
  journal = {arXiv preprint arXiv:2505.09388},
  year = {2025}
}

@article{zhang2026atod,
  title={{ATOD}: An Evaluation Framework and Benchmark for Agentic Task-Oriented Dialogue Systems},
  author={Zhang, Yifei and Nayyeri, Hooshang and Khaziev, Rinat and Yilmaz, Emine and Tur, Gokhan and Hakkani-T{\"u}r, Dilek and Thadakamalla, Hari},
  journal={arXiv preprint arXiv:2601.11854},
  year={2026}
}

@article{young2013pomdp,
  title={POMDP-based statistical spoken dialog systems: A review},
  author={Young, Steve and Ga{\v{s}}i{\'c}, Milica and Thomson, Blaise and Williams, Jason D.},
  journal={Proceedings of the IEEE},
  volume={101},
  number={5},
  pages={1160--1179},
  year={2013},
  publisher={IEEE}
}

@misc{zhou2026voxcpm2,
  title = {{VoxCPM2 Technical Report}},
  author = {Zhou, Yixuan and Zeng, Guoyang and Liu, Xin and Li, Xiang and Yu, Renjie and Gui, Jiancheng and Wu, Jiaheng and Wang, Ziyang and Shen, Xudong and Ye, Runchuan and Zhang, Zhisheng and Zhou, Jiuyang and Bai, Bingsong and Sun, Weiyue and Deng, Mengyuan and Shi, Qundong and Wu, Zhiyong and Liu, Zhiyuan},
  year = {2026},
  eprint = {2606.06928},
  archivePrefix = {arXiv},
  primaryClass = {eess.AS}
}

@misc{hu2026qwen3tts,
  title = {{Qwen3-TTS Technical Report}},
  author = {Hu, Hangrui and Zhu, Xinfa and He, Ting and Guo, Dake and Zhang, Bin and Wang, Xiong and Guo, Zhifang and Jiang, Ziyue and Hao, Hongkun and Guo, Zishan and Zhang, Xinyu and Zhang, Pei and Yang, Baosong and Xu, Jin and Zhou, Jingren and Lin, Junyang},
  year = {2026},
  eprint = {2601.15621},
  archivePrefix = {arXiv},
  primaryClass = {cs.SD}
}

@inproceedings{si2023spokenwoz,
  title = {{SpokenWOZ}: A Large-Scale Speech-Text Benchmark for Spoken Task-Oriented Dialogue Agents},
  author = {Si, Shuzheng and Ma, Wentao and Gao, Haoyu and Wu, Yuchuan and Lin, Ting-En and Dai, Yinpei and Li, Hangyu and Yan, Rui and Huang, Fei and Li, Yongbin},
  booktitle = {Advances in Neural Information Processing Systems},
  volume = {36},
  year = {2023}
}

@inproceedings{ao2024sdeval,
  title = {{SD-Eval}: A Benchmark Dataset for Spoken Dialogue Understanding Beyond Words},
  author = {Ao, Junyi and Wang, Yuancheng and Tian, Xiaohai and Chen, Dekun and Zhang, Jun and Lu, Lu and Wang, Yuxuan and Li, Haizhou and Wu, Zhizheng},
  booktitle = {Advances in Neural Information Processing Systems},
  volume = {37},
  year = {2024}
}

@inproceedings{yang2021superb,
  title = {{SUPERB}: Speech Processing Universal PERformance Benchmark},
  author = {Yang, Shu-wen and Chi, Po-Han and Chuang, Yung-Sung and Lai, Cheng-I Jeff and Lakhotia, Kushal and Lin, Yist Y. and Liu, Andy T. and Shi, Jiatong and Chang, Xuankai and Lin, Guan-Ting and Huang, Tzu-Hsien and Tseng, Wei-Cheng and Lee, Ko-tik and Liu, Da-Rong and Huang, Zili and Dong, Shuyan and Li, Shang-Wen and Watanabe, Shinji and Mohamed, Abdelrahman and Lee, Hung-yi},
  booktitle = {Interspeech 2021},
  pages = {1194--1198},
  year = {2021},
  doi = {10.21437/Interspeech.2021-1775},
  url = {https://www.isca-archive.org/interspeech_2021/yang21c_interspeech.html}
}

@misc{qwen25omni2025,
  title = {{Qwen2.5-Omni} Technical Report},
  author = {Xu, Jin and Guo, Zhifang and He, Jinzheng and Hu, Hangrui and He, Ting and Bai, Shuai and Chen, Keqin and Wang, Jialin and Fan, Yang and Dang, Kai and Zhang, Bin and Wang, Xiong and Chu, Yunfei and Lin, Junyang},
  year = {2025},
  eprint = {2503.20215},
  archivePrefix = {arXiv},
  primaryClass = {cs.CL},
  url = {https://arxiv.org/abs/2503.20215}
}

@inproceedings{budzianowski2018multiwoz,
  title={{MultiWOZ} -- A Large-Scale Multi-Domain Wizard-of-Oz Dataset for Task-Oriented Dialogue Modelling},
  author={Budzianowski, Pawe{\l} and Wen, Tsung-Hsien and Tseng, Bo-Hsiang and Casanueva, I{\~n}igo and Ultes, Stefan and Ramadan, Osman and Ga{\v{s}}i{\'c}, Milica},
  booktitle={Proceedings of the 2018 Conference on Empirical Methods in Natural Language Processing (EMNLP)},
  pages={5016--5026},
  year={2018}
}

@inproceedings{feng2022emowoz,
  title={{EmoWOZ}: A Large-Scale Corpus and Labelling Scheme for Emotion Recognition in Task-Oriented Dialogue Systems},
  author={Feng, Shutong and Lubis, Nurul and Geishauser, Christian and Lin, Hsien-Chin and Heck, Michael and van Niekerk, Carel and Ga{\v{s}}i{\'c}, Milica},
  booktitle={Proceedings of the Thirteenth Language Resources and Evaluation Conference (LREC)},
  pages={4096--4113},
  year={2022}
}

@inproceedings{lin2023emous,
  title={{EmoUS}: Simulating User Emotions in Task-Oriented Dialogues},
  author={Lin, Hsien-Chin and Feng, Shutong and Geishauser, Christian and Lubis, Nurul and van Niekerk, Carel and Heck, Michael and Ruppik, Benjamin and Vukovic, Renato and Ga{\v{s}}i{\'c}, Milica},
  booktitle={Proceedings of the 46th International ACM SIGIR Conference on Research and Development in Information Retrieval (SIGIR)},
  pages={2526--2531},
  year={2023}
}

@inproceedings{feng2024emoloop,
  title = {Infusing Emotions into Task-oriented Dialogue Systems: Understanding, Management, and Generation},
  author = {Feng, Shutong and Lin, Hsien-Chin and Geishauser, Christian and Lubis, Nurul and van Niekerk, Carel and Heck, Michael and Ruppik, Benjamin and Vukovic, Renato and Ga{\v{s}}i{\'c}, Milica},
  booktitle = {Proceedings of the 25th Annual Meeting of the Special Interest Group on Discourse and Dialogue},
  pages = {699--717},
  year = {2024},
  address = {Kyoto, Japan},
  publisher = {Association for Computational Linguistics},
  doi = {10.18653/v1/2024.sigdial-1.60},
  url = {https://aclanthology.org/2024.sigdial-1.60/}
}

@inproceedings{yang2024airbench,
  title = {{AIR-Bench}: Benchmarking Large Audio-Language Models via Generative Comprehension},
  author = {Yang, Qian and Xu, Jin and Liu, Wenrui and Chu, Yunfei and Jiang, Ziyue and Zhou, Xiaohuan and Leng, Yichong and Lv, Yuanjun and Zhao, Zhou and Zhou, Chang and Zhou, Jingren},
  booktitle = {Proceedings of the 62nd Annual Meeting of the Association for Computational Linguistics (Volume 1: Long Papers)},
  pages = {1979--1998},
  year = {2024},
  address = {Bangkok, Thailand},
  publisher = {Association for Computational Linguistics},
  doi = {10.18653/v1/2024.acl-long.109},
  url = {https://aclanthology.org/2024.acl-long.109/}
}

@inproceedings{huang2024dynamicsuperb,
  title = {Dynamic-{SUPERB}: Towards a Dynamic, Collaborative, and Comprehensive Instruction-Tuning Benchmark for Speech},
  author = {Huang, Chien-yu and Lu, Ke-Han and Wang, Shih-Heng and Hsiao, Chi-Yuan and Kuan, Chun-Yi and Wu, Haibin and Arora, Siddhant and Chang, Kai-Wei and Shi, Jiatong and Peng, Yifan and Sharma, Roshan and Watanabe, Shinji and Ramakrishnan, Bhiksha and Shehata, Shady and Lee, Hung-yi},
  booktitle = {2024 IEEE International Conference on Acoustics, Speech and Signal Processing (ICASSP)},
  pages = {12136--12140},
  year = {2024},
  doi = {10.1109/ICASSP48485.2024.10448257}
}

@inproceedings{wang2025audiobench,
  title = {{AudioBench}: A Universal Benchmark for Audio Large Language Models},
  author = {Wang, Bin and Zou, Xunlong and Lin, Geyu and Sun, Shuo and Liu, Zhuohan and Zhang, Wenyu and Liu, Zhengyuan and Aw, AiTi and Chen, Nancy F.},
  booktitle = {Proceedings of the 2025 Conference of the Nations of the Americas Chapter of the Association for Computational Linguistics: Human Language Technologies (Volume 1: Long Papers)},
  pages = {4297--4316},
  year = {2025},
  address = {Albuquerque, New Mexico},
  publisher = {Association for Computational Linguistics},
  doi = {10.18653/v1/2025.naacl-long.218},
  url = {https://aclanthology.org/2025.naacl-long.218/}
}

@misc{chu2024qwen2audio,
  title = {{Qwen2-Audio} Technical Report},
  author = {Chu, Yunfei and Xu, Jin and Yang, Qian and Wei, Haojie and Wei, Xipin and Guo, Zhifang and Leng, Yichong and Lv, Yuanjun and He, Jinzheng and Lin, Junyang and Zhou, Chang and Zhou, Jingren},
  year = {2024},
  eprint = {2407.10759},
  archivePrefix = {arXiv},
  primaryClass = {cs.CL},
  url = {https://arxiv.org/abs/2407.10759}
}

@inproceedings{rastogi2020sgd,
  title={Towards Scalable Multi-domain Conversational Agents: The Schema-Guided Dialogue Dataset},
  author={Rastogi, Abhinav and Zang, Xiaoxue and Sunkara, Srinivas and Gupta, Raghav and Khaitan, Pranav},
  booktitle={Proceedings of the AAAI Conference on Artificial Intelligence},
  volume={34},
  pages={8689--8696},
  year={2020}
}

@inproceedings{lin2024styletalk,
  title = {Advancing Large Language Models to Capture Varied Speaking Styles and Respond Properly in Spoken Conversations},
  author = {Lin, Guan-Ting and Chiang, Cheng-Han and Lee, Hung-yi},
  booktitle = {Proceedings of the 62nd Annual Meeting of the Association for Computational Linguistics (Volume 1: Long Papers)},
  pages = {6626--6642},
  year = {2024},
  address = {Bangkok, Thailand},
  publisher = {Association for Computational Linguistics},
  doi = {10.18653/v1/2024.acl-long.358},
  url = {https://aclanthology.org/2024.acl-long.358/}
}

@inproceedings{paras2s2025,
  title={{ParaS2S}: Benchmarking and Aligning Spoken Language Models for Paralinguistic-aware Speech-to-Speech Interaction},
  author={Yang, Shu-wen and Tu, Ming and Liu, Ting-Wei and Qu, Xinghua and Lee, Hung-yi and Lu, Lu and Wang, Yuxuan and Wu, Yonghui},
  booktitle={International Conference on Learning Representations},
  year={2026},
  url={https://proceedings.iclr.cc/paper_files/paper/2026/hash/734b2e77222680728b9ce78c573eae1e-Abstract-Conference.html}
}

@inproceedings{gao2024adubench,
  title = {Benchmarking Open-ended Audio Dialogue Understanding for Large Audio-Language Models},
  author = {Gao, Kuofeng and Xia, Shu-Tao and Xu, Ke and Torr, Philip and Gu, Jindong},
  booktitle = {Proceedings of the 63rd Annual Meeting of the Association for Computational Linguistics (Volume 1: Long Papers)},
  pages = {4763--4784},
  year = {2025},
  address = {Vienna, Austria},
  publisher = {Association for Computational Linguistics},
  doi = {10.18653/v1/2025.acl-long.237},
  url = {https://aclanthology.org/2025.acl-long.237/}
}

@misc{multibench2025,
  title={{MULTI-Bench}: A Multi-Turn Interactive Benchmark for Assessing Emotional Intelligence Ability of Spoken Dialogue Models},
  author={Deng, Yayue and Hu, Guoqiang and Sun, Haiyang and Zhang, Xiangyu and Zhang, Haoyang and Tian, Fei and Yang, Xuerui and Yu, Gang and Chng, Eng Siong},
  howpublished={arXiv:2511.00850},
  year={2025}
}

@misc{humdial2026,
  title={{HumDial-EIBench}: A Human-Recorded Multi-Turn Emotional Intelligence Benchmark for Audio Language Models},
  author={Wang, Shuiyuan and Zhao, Zhixian and Xue, Hongfei and Wang, Chengyou and Wang, Shuai and Bu, Hui and Xu, Xin and Xie, Lei},
  howpublished={arXiv:2604.11594},
  year={2026}
}

@inproceedings{realtalkcn2025,
  title = {{RealTalk-CN}: A Realistic Chinese Speech Task-Oriented Dialogue Benchmark with Cross-Modal Analysis},
  author = {Wang, Enzhi and Zhou, Jiaming and Jia, Yuhang and Kong, Aobo and Li, Qicheng and Qin, Yong},
  booktitle = {Proceedings of the 64th Annual Meeting of the Association for Computational Linguistics (Volume 1: Long Papers)},
  pages = {2880--2897},
  year = {2026},
  address = {San Diego, California, United States},
  publisher = {Association for Computational Linguistics},
  doi = {10.18653/v1/2026.acl-long.131},
  url = {https://aclanthology.org/2026.acl-long.131/}
}

@misc{vocalaffectbench2026,
  title = {{VocalAffectBench: Evaluating Vocal Emotion Recognition in AI Audio Models}},
  author = {Debaupte, Luc and Baumgartner, Tyler and Tai, Brandon and Fan, Candice and Wang, Bill and Zhong, Yi},
  year = {2026},
  howpublished = {Hugging Face dataset, \url{https://huggingface.co/datasets/besimple-ai/vocal-affect-bench}}
}

@inproceedings{chen2026listen,
  title = {Do Audio {LLM}s Really {LISTEN}, or Just Transcribe? Measuring Lexical vs. Acoustic Emotion Cues Reliance},
  author = {Chen, Jingyi and Guo, Zhimeng and Chun, Jiyun and Wang, Pichao and Perrault, Andrew and Elsner, Micha},
  booktitle = {Proceedings of the 19th Conference of the European Chapter of the Association for Computational Linguistics (Volume 1: Long Papers)},
  pages = {5848--5877},
  address = {Rabat, Morocco},
  publisher = {Association for Computational Linguistics},
  year = {2026},
  month = mar,
  doi = {10.18653/v1/2026.eacl-long.274},
  url = {https://aclanthology.org/2026.eacl-long.274/}
}

@inproceedings{kim2026aligning,
  title = {Aligning Paralinguistic Understanding and Generation in Speech {LLM}s via Multi-Task Reinforcement Learning},
  author = {Kim, Minseok and Chen, Jingxiang and Leem, Seong-Gyun and Huang, Yin and Rungta, Rashi and Ouyang, Zhicheng and Wu, Haibin and Appini, Surya Teja and Bansal, Ankur and Bai, Yang and Liu, Yue and Metze, Florian and Aly, Ahmed A and Kumar, Anuj and Rastrow, Ariya and Lin, Zhaojiang},
  booktitle = {Proceedings of the 19th Conference of the European Chapter of the Association for Computational Linguistics (Volume 5: Industry Track)},
  pages = {636--648},
  address = {Rabat, Morocco},
  publisher = {Association for Computational Linguistics},
  year = {2026},
  month = mar,
  doi = {10.18653/v1/2026.eacl-industry.49},
  url = {https://aclanthology.org/2026.eacl-industry.49/}
}

@article{wang2026parabridge,
  title = {{ParaBridge}: Bridging Paralinguistic Perception and Dialogue Behavior in Speech Language Models},
  author = {Wang, Yuxiang and Ni, Qinke and Cai, Shengbo and Lin, Wan and Zhang, Liqiang and Wu, Zhizheng},
  journal = {arXiv preprint arXiv:2606.10581},
  year = {2026},
  doi = {10.48550/arXiv.2606.10581},
  url = {https://arxiv.org/abs/2606.10581}
}

@inproceedings{miyazawa2026paralinguistic,
  title = {Evaluation of Paralinguistic-Aware Spoken Dialogue Systems using Next-Utterance Classification},
  author = {Miyazawa, Kouki and Sato, Yoshinao},
  booktitle = {Proceedings of the 27th Annual Meeting of the Special Interest Group on Discourse and Dialogue},
  pages = {711--719},
  address = {Atlanta, Georgia, USA},
  publisher = {Association for Computational Linguistics},
  year = {2026},
  month = aug,
  url = {https://aclanthology.org/2026.sigdial-1.50/}
}

@article{hsu2021hubert,
  title = {{HuBERT}: Self-Supervised Speech Representation Learning by Masked Prediction of Hidden Units},
  author = {Hsu, Wei-Ning and Bolte, Benjamin and Tsai, Yao-Hung Hubert and Lakhotia, Kushal and Salakhutdinov, Ruslan and Mohamed, Abdelrahman},
  journal = {IEEE/ACM Transactions on Audio, Speech, and Language Processing},
  volume = {29},
  pages = {3451--3460},
  year = {2021},
  doi = {10.1109/TASLP.2021.3122291}
}

@article{ravanelli2021speechbrain,
  title = {{SpeechBrain}: A General-Purpose Speech Toolkit},
  author = {Ravanelli, Mirco and Parcollet, Titouan and Plantinga, Peter and Rouhe, Aku and Cornell, Samuele and Lugosch, Loren and Subakan, Cem and Dawalatabad, Nauman and Heba, Abdelwahab and Zhong, Jianyuan and Chou, Ju-Chieh and Yeh, Sung-Lin and Fu, Szu-Wei and Liao, Chien-Feng and Rastorgueva, Elena and Grondin, Fran{\c{c}}ois and Aris, William and Na, Hwidong and Gao, Yan and De Mori, Renato and Bengio, Yoshua},
  journal = {arXiv preprint arXiv:2106.04624},
  year = {2021},
  url = {https://arxiv.org/abs/2106.04624}
}

@inproceedings{baevski2020wav2vec2,
  title = {wav2vec 2.0: A Framework for Self-Supervised Learning of Speech Representations},
  author = {Baevski, Alexei and Zhou, Yuhao and Mohamed, Abdelrahman and Auli, Michael},
  booktitle = {Advances in Neural Information Processing Systems (NeurIPS)},
  volume = {33},
  pages = {12449--12460},
  year = {2020},
  url = {https://proceedings.neurips.cc/paper/2020/hash/92d1e1eb1cd6f9fba3227870bb6d7f07-Abstract.html}
}

\appendix

\section{SGD-Seeded Domains and Subdomain Scenarios}
\label{app:domains}

Table~\ref{tab:domains} shows the complete domain inventory: 48 consumer-service domains seeded from the service categories of the Schema-Guided Dialogue dataset and grouped into eight service families. Rather than listing all 480 subdomain seeds, Table~\ref{tab:domainexample} expands one domain in full---the budget-airline domain used in Figure~\ref{fig:framework}---to illustrate how scenario seeds vary in user situation and decision pressure. The resulting variation in urgency, budget tier, expertise, and life context grounds the scenario-specific hidden ladders described in Section~\ref{sec:scenario}; the construction procedure is summarized in Appendix~\ref{app:prompts:gen}.

\begin{table*}[!t]
\centering
\caption{\textbf{Domain inventory.}
Hear2Act covers 48 consumer-service domains grouped into eight families.
Each domain is expanded into ten scenario seeds, yielding 480 benchmark scenarios.}
\label{tab:domains}

\vspace{2pt}
\footnotesize
\setlength{\tabcolsep}{4pt}
\renewcommand{\arraystretch}{1.16}

\begin{tabularx}{\textwidth}{
    @{}
    >{\raggedright\arraybackslash}p{0.17\textwidth}
    >{\centering\arraybackslash}p{0.035\textwidth}
    >{\raggedright\arraybackslash}X
    @{}
}
\toprule
\textbf{Service family} & \textbf{\#} & \textbf{Domains} \\
\midrule

\textbf{Travel \& Events}
& 5
& Budget airline tickets, vacation rental properties, restaurant reservations,
concert ticket purchases, and wedding venue selection. \\

\addlinespace[2pt]
\textbf{Home \& Property}
& 8
& Home cleaning services, home renovation contractors, home security systems,
landscaping contractors, lawn care contractors, solar panel installation,
kitchen appliance upgrades, and mattress replacement. \\

\addlinespace[2pt]
\textbf{Finance \& Insurance}
& 9
& Credit card applications, mortgage lender comparison, investment portfolio
allocation, retirement planning advisors, tax preparation services, car insurance
policies, health insurance plans, pet insurance policies, and business insurance
coverage. \\

\addlinespace[2pt]
\textbf{Health \& Wellness}
& 7
& Dermatologist appointments, pediatrician selection, mental health therapists,
meditation apps, gym membership options, fitness tracker devices, and prescription
eyeglasses. \\

\addlinespace[2pt]
\textbf{Education \& Career}
& 5
& College major selection, online coding bootcamps, language learning platforms,
professional development courses, and laptop purchase for students. \\

\addlinespace[2pt]
\textbf{Family \& Lifestyle}
& 6
& Children's daycare centers, dog training classes, online dating platforms,
wine club memberships, meal delivery subscriptions, and video game purchases. \\

\addlinespace[2pt]
\textbf{Media \& Devices}
& 4
& Cable TV packages, streaming service subscriptions, podcast hosting services,
and smartphone upgrades. \\

\addlinespace[2pt]
\textbf{Professional Services}
& 4
& Legal consultation services, auto mechanic services, business accounting
software, and freelance graphic designers. \\

\midrule
\textbf{Total}
& \textbf{48}
& \textbf{480 scenarios across ten seeds per domain} \\
\bottomrule
\end{tabularx}
\end{table*}
\begin{table*}[!t]
\centering
\caption{\textbf{One expanded domain.}
 Ten subdomain seeds for budget airline tickets illustrate variation in user situation and decision pressure before scenario instantiation.}
\label{tab:domainexample}

\vspace{2pt}
\small
\setlength{\tabcolsep}{7pt}
\renewcommand{\arraystretch}{1.22}

\begin{tabular}{@{}c
>{\raggedright\arraybackslash}p{0.34\textwidth}
>{\raggedright\arraybackslash}p{0.55\textwidth}@{}}
\toprule
& \textbf{Subdomain situation (who / pressure)}
& \textbf{Opening request} \\
\midrule

1 & Last-minute emergency travel for family medical situation with extremely limited budget
& {\itshape``I need to fly to see my sick grandmother tomorrow but only have \$200---what are my cheapest options?''} \\

2 & College student planning spring break trip with friends on tight budget
& {\itshape``Can you help me find the cheapest flights for four college students going to Miami for spring break?''} \\

3 & Budget-conscious family of five planning annual vacation
& {\itshape``What's the most affordable way to fly my family of five to Orlando for our Disney World trip?''} \\

4 & Digital nomad seeking flexible travel dates for extended European backpacking
& {\itshape``I want to backpack through Europe for 3 months---which budget airlines offer the best multi-city deals?''} \\

5 & Job interview candidate needing quick affordable travel for an unexpected opportunity
& {\itshape``I have a job interview in Seattle next week and need the cheapest flight possible from Chicago.''} \\

6 & Retiree on fixed income wanting to visit grandchildren regularly
& {\itshape``As a senior on a fixed income, what budget airline options exist for regular visits to see my grandkids?''} \\

7 & Young professional attending a destination wedding with multiple flight segments
& {\itshape``I need budget flights to get to my friend's wedding in Bali, including connections---what's the cheapest route?''} \\

8 & Small business owner traveling frequently for client meetings on a startup budget
& {\itshape``I need to travel monthly for business but my startup has a tight travel budget---which airlines offer the best deals for frequent short trips?''} \\

9 & International student trying to visit home during semester break
& {\itshape``I'm an international student wanting to fly home to India for winter break---what are the most affordable long-haul options?''} \\

10 & Adventure traveler planning a multi-stop trip to remote destinations
& {\itshape``I want to visit three different countries in South America on a backpacker's budget---which budget airlines serve those routes?''} \\

\bottomrule
\end{tabular}
\end{table*}

\section{VoxCPM2 Renderer Replication}
\label{app:audiogrid}

\begin{table*}[!t]
\centering
\caption{\textbf{Full spoken-assistant condition grid for Qwen2.5-Omni} with VoxCPM2 ($n{=}480$ per condition).
$T$, $A$, $S$, and $\hat S$ denote transcript, audio, ground-truth state, and audio-inferred state. Audio-derived representations are supplied as text alongside the transcript. Bold/underline mark the highest/next-highest value per column.}
\label{tab:audioidx}

\vspace{2pt}
\small
\setlength{\tabcolsep}{2.8pt}
\renewcommand{\arraystretch}{1.06}

\begin{tabular}{
@{}
>{\raggedright\arraybackslash}p{0.31\textwidth}
@{\hspace{5pt}}
rrr
@{\hspace{22pt}}
rrr
@{}
}
\toprule
& \multicolumn{3}{c}{\textbf{Prosody-mediated responses}}
& \multicolumn{3}{c}{\textbf{Explicit lexical responses}} \\

\cmidrule(l{5pt}r{5pt}){2-4}
\cmidrule(l{5pt}r{5pt}){5-7}

\textbf{Input / representation}
& \shortstack{\strut Success\\\%$\uparrow$}
& \shortstack{\strut Accept\\\%$\uparrow$}
& \shortstack{\strut FullFit\\\%$\uparrow$}
& \shortstack{\strut Success\\\%$\uparrow$}
& \shortstack{\strut Accept\\\%$\uparrow$}
& \shortstack{\strut FullFit\\\%$\uparrow$} \\
\midrule

\multicolumn{7}{@{}l}{\emph{Direct input}} \\[0.5pt]

\quad Transcript only ($T$)
& 11.9 & 22.5 & 11.7
& \underline{52.7} & 44.6 & \underline{28.9} \\

\quad Audio only ($A$)
& 15.4 & 22.6 & 13.8
& \textbf{53.3} & \textbf{45.7} & 28.2 \\

\quad Audio + transcript ($A+T$)
& 16.2 & 25.6 & 14.6
& 48.3 & 45.1 & 28.3 \\

\addlinespace[1.5pt]
\multicolumn{7}{@{}l}{\emph{Textualized prosodic representations}} \\[0.5pt]

\quad Generic affect (HuBERT)
& 35.2 & 40.2 & 24.0
& 51.7 & 45.0 & 27.8 \\

\quad Generic affect (SpeechBrain)
& 32.1 & 40.0 & 24.1
& 51.5 & 45.0 & 27.8 \\

\quad Audio-inferred state ($T+\hat S$)
& \underline{42.1} & \underline{42.4} & \underline{25.6}
& 51.5 & \underline{45.2} & 28.7 \\

\addlinespace[0.5pt]
\midrule

\emph{Ground-truth state ($T+S$)}
& \textbf{\textit{45.4}}
& \textbf{\textit{42.5}}
& \textbf{\textit{27.5}}
& \textit{52.1}
& \textbf{\textit{45.7}}
& \textbf{\textit{30.1}} \\

\bottomrule
\end{tabular}
\end{table*}

Table~\ref{tab:audioidx} repeats the Qwen2.5-Omni condition grid with all speech re-rendered by VoxCPM2. The qualitative pattern of Section~\ref{sec:rq3} is preserved: raw audio tracks the transcript-only baseline and the audio-inferred-state condition recovers the text-side gain. Under Prosody-mediated responses, the two SER-derived conditions also keep their position, each matching its Qwen3-TTS value to one decimal on Success\% and both still below the audio-inferred state on all three metrics. Across the $42$ condition cells the per-cell difference from the Qwen3-TTS arm (Tables~\ref{tab:audio} and~\ref{tab:audioexp}) is at most $3.5$ points and $1.4$ on average---the pattern is therefore not specific to one speech renderer.

\section{Additional Text-Model Behavior Analyses}
\label{app:textbehavior}

\begin{figure*}[!t]
\centering
\includegraphics[width=\textwidth]{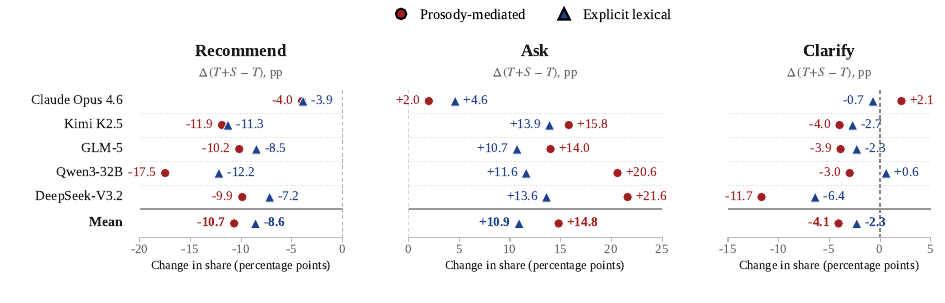}
\caption{\textbf{Change in assistant action composition with state access.}
Points show the $T{+}S{-}T$ change, in percentage points, in the percentage of assistant decision turns assigned to \textsc{recommend}, \textsc{ask}, and \textsc{clarify} under Prosody-mediated and Explicit lexical responses. Means are macro-averaged across models.}
\label{fig:actioncomp}
\end{figure*}

Figure~\ref{fig:actioncomp} shows that state access shifts assistant behavior away from early recommendation and toward information gathering. The percentage of \textsc{recommend} turns falls and the percentage of \textsc{ask} turns rises for every model, with larger changes under Prosody-mediated responses. Changes in \textsc{clarify} vary across models.

\begin{figure}[!t]
\centering
\includegraphics[width=\columnwidth]{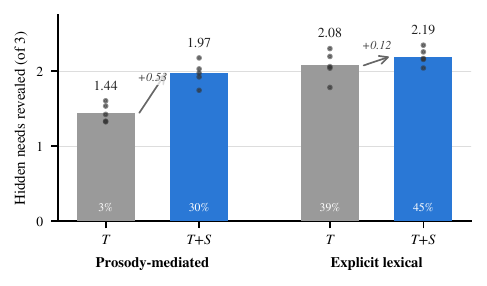}
\caption{\textbf{State access increases hidden-need disclosure.}
Bars show the mean number of hidden needs revealed, dots show per-model means, and in-bar percentages show the proportion of rollouts in which all three hidden needs are revealed ($T$: transcript only; $T{+}S$: transcript plus turn-level ground-truth state). The effect is larger under Prosody-mediated responses, where the relevant preferences remain lexically implicit.}
\label{fig:disclosure}
\end{figure}

Figure~\ref{fig:disclosure} shows a corresponding increase in hidden-need disclosure. Under Prosody-mediated responses, the mean number of revealed needs rises from 1.44 to 1.97, and full disclosure rises from 3\% to 30\%. The changes are smaller under Explicit lexical responses, where much of the preference information is already available in the transcript.

\paragraph{Residual gain under Explicit lexical responses.}
The remaining task-success gain from state access is concentrated in L3 turns expressed as qualified assent. After such turns, immediate \textsc{confirm} falls from 17.2\% with $T$ to 1.7\% with $T{+}S$, while \textsc{ask} rises from 25.9\% to 70.2\%. The rate of final selections that violate only L3 also falls from 19.5\% to 9.1\%. In contrast, after turns that directly state an L2 preference, immediate confirmation is already rare, at 1.9\% with $T$ and 0.6\% with $T{+}S$. These results support the interpretation that state access is most useful under Explicit lexical responses when the preference is verbalized but the utterance still has an accepting surface form.

\section{Confidence Intervals for Key Diagnostic Contrasts}
\label{app:ci}

Table~\ref{tab:ci} reports scenario-level paired bootstrap 95\% confidence intervals for selected diagnostic contrasts used in the analysis: the pooled text-LLM state-access effects of Section~\ref{sec:rq1}, matched spoken-assistant state comparisons from Section~\ref{sec:rq3}, and successive steps of the label-fidelity intervention in Section~\ref{sec:rq2}. Scenarios are resampled jointly across paired conditions, so each interval reflects scenario-level variation under the pairing that produced the reported point estimate.

\begin{table*}[!t]
\centering
\caption{\textbf{Scenario-level paired bootstrap 95\% CIs for key diagnostic contrasts.}
CIs use 2{,}000 joint scenario resamples. Point estimates correspond to
Tables~\ref{tab:text}, \ref{tab:audio}, and~\ref{tab:audioexp}, and
Figure~\ref{fig:failure}. $T$, $S$, and $\hat S$ denote transcript,
ground-truth state, and audio-inferred state, respectively.
The label-fidelity analysis uses 48 intervention scenarios.}
\label{tab:ci}

\vspace{2pt}
\small
\setlength{\tabcolsep}{8pt}
\renewcommand{\arraystretch}{1.18}

\resizebox{0.78\textwidth}{!}{%
\begin{tabular}{@{}l rr@{}}
\toprule
\textbf{Contrast} & \emph{Prosody-mediated} & \emph{Explicit lexical} \\
\midrule

\multicolumn{3}{@{}l}{\emph{Text LLMs, pooled over five models
($\Delta=(T+S)-T$)}}\\[1pt]

\quad Success\%
& +36.8\;{\color{black!55}\footnotesize[+34.4,\,+39.1]}
& +6.3\;{\color{black!55}\footnotesize[+4.6,\,+8.0]} \\

\quad Accept\%
& +7.5\;{\color{black!55}\footnotesize[+6.0,\,+8.9]}
& $-$1.7\;{\color{black!55}\footnotesize[$-$2.6,\,$-$0.8]} \\

\quad FullFit\%
& +4.1\;{\color{black!55}\footnotesize[+2.9,\,+5.4]}
& +0.3\;{\color{black!55}\footnotesize[$-$0.3,\,+0.9]} \\

\addlinespace[4pt]
\multicolumn{3}{@{}l}{\emph{Qwen2.5-Omni-7B, Success\% diagnostic contrasts}}\\[1pt]

\quad Ground-truth state vs.\ transcript ($(T+S)-T$)
& +29.2\;{\color{black!55}\footnotesize[+24.0,\,+34.6]}
& $-$6.7\;{\color{black!55}\footnotesize[$-$13.1,\,$-$0.4]} \\

\quad Audio-inferred vs.\ ground-truth state ($(T+\hat S)-(T+S)$)
& $-$1.5\;{\color{black!55}\footnotesize[$-$7.3,\,+4.6]}
& +2.7\;{\color{black!55}\footnotesize[$-$3.5,\,+9.0]} \\

\addlinespace[4pt]
\multicolumn{3}{@{}l}{\emph{Qwen2-Audio-7B-Instruct, Success\% diagnostic contrasts}}\\[1pt]

\quad Ground-truth state vs.\ transcript ($(T+S)-T$)
& +23.1\;{\color{black!55}\footnotesize[+20.0,\,+26.2]}
& +2.1\;{\color{black!55}\footnotesize[$-$1.7,\,+5.9]} \\

\quad Audio-inferred vs.\ ground-truth state ($(T+\hat S)-(T+S)$)
& $-$0.2\;{\color{black!55}\footnotesize[$-$3.8,\,+3.5]}
& $-$0.4\;{\color{black!55}\footnotesize[$-$4.3,\,+3.4]} \\

\addlinespace[4pt]
\multicolumn{3}{@{}l}{\emph{Label-fidelity ladder, Success\% successive steps
(48 scenarios, pooled models)}}\\[1pt]

\quad Always resolved $-$ no state
& +0.6\;{\color{black!55}\footnotesize[$-$4.4,\,+5.4]}
& $-$5.0\;{\color{black!55}\footnotesize[$-$11.0,\,+0.2]} \\

\quad Shuffled $-$ always resolved
& +7.1\;{\color{black!55}\footnotesize[+1.2,\,+12.9]}
& +3.3\;{\color{black!55}\footnotesize[$-$2.9,\,+9.6]} \\

\quad Always unresolved $-$ shuffled
& +10.8\;{\color{black!55}\footnotesize[+4.2,\,+17.9]}
& +2.1\;{\color{black!55}\footnotesize[$-$3.3,\,+7.5]} \\

\quad Correct state $-$ always unresolved
& +14.8\;{\color{black!55}\footnotesize[+7.9,\,+21.7]}
& +5.2\;{\color{black!55}\footnotesize[+0.4,\,+10.4]} \\

\bottomrule
\end{tabular}%
}
\end{table*}

\section{State-Cue Perception Human Study}
\label{app:tonestudy}

This appendix provides details of the human perception study reported in Section~\ref{sec:rq3}, including the full speech-rendering validation results in Table~\ref{tab:human}.

\begin{table}[!htb]
\centering
\caption{\textbf{Speech-rendering validation.}
Human listeners recover resolved versus unresolved state from both renderers with $0.92$ accuracy ($n{=}100$ per renderer; balanced classes), confirming that the intended prosodic contrast remains perceptible after rendering.
$\kappa$ denotes inter-annotator agreement.}
\label{tab:human}
\vspace{3pt}
\renewcommand{\arraystretch}{1.15}\small
\begin{tabular}{@{}lcc@{}}
\toprule
State & Qwen3-TTS & VoxCPM2 \\
\midrule
Resolved ($O^{+}$) & 0.93 & 0.90 \\
Unresolved ($O^{-}$) & 0.91 & 0.93 \\
\midrule
Overall & 0.92 & 0.92 \\
$\kappa$ (annotators) & 0.96 & 0.90 \\
\bottomrule
\end{tabular}
\end{table}

\paragraph{Participants and materials.} Two graduate-student annotators based in the United States independently annotated the synthetic speech clips and were compensated at \$25 per hour. Participation was voluntary. For each renderer, \HToneN{} utterances were sampled from the rollout set, balanced $50/50$ over the intended collapsed state (\textsc{Resolved} vs.\ \textsc{Unresolved}). Each item included the preceding dialogue context and one rendered user utterance. No real-user data, personal information, or sensitive information was presented or collected. No additional annotator demographics were collected, and the results are not intended to support population-level claims about human perception.

\paragraph{Annotation task.} Annotators judged the resolved/unresolved state conveyed by each rendered utterance. They used the dialogue context to identify the recommendation under discussion, but were instructed not to infer the user's hidden needs, identity, personality, demographic characteristics, or clinical or psychological state. Qwen3-TTS clips used a four-way choice over the benchmark's graded states, which was collapsed to \textsc{Resolved} versus \textsc{Unresolved} for the reported analysis. VoxCPM2 clips used the binary distinction directly. The primary spoken assistant, Qwen2.5-Omni-7B, was evaluated on the same clips with the same dialogue context and corresponding label space.

\paragraph{Qwen3-TTS annotation protocol.} The four-way task used the following instructions:

\begin{quote}
\small
You will hear a synthetic user utterance together with the dialogue context that precedes it. Judge how the speaker responds to the assistant's current recommendation based on the rendered utterance, especially its vocal delivery. Select exactly one label.

\textbf{Genuine acceptance}: The speaker sounds satisfied and treats the current recommendation as resolved.

\textbf{Reluctant acceptance}: The words may appear accepting, but the speaker sounds hesitant, lukewarm, underwhelmed, or otherwise unconvinced. The current recommendation remains unresolved.

\textbf{Voiced concern}: The speaker raises or clearly conveys a concern about the current recommendation. The current recommendation remains unresolved.

\textbf{Rejection}: The speaker clearly rejects the current recommendation or conveys strong dissatisfaction with it. The current recommendation remains unresolved.

Use the dialogue context only to identify what the speaker is responding to. Do not guess the speaker's hidden preferences, identity, personality, demographic characteristics, or clinical or psychological state. If more than one label appears plausible, choose the label that best captures whether the speaker treats the recommendation as resolved and, if not, how strongly the utterance signals otherwise.
\end{quote}

\paragraph{VoxCPM2 annotation protocol.} The binary task used the following instructions:

\begin{quote}
\small
You will hear a synthetic user utterance together with the dialogue context that precedes it. Judge whether the speaker treats the assistant's current recommendation as resolved or unresolved based on the rendered utterance, especially its vocal delivery. Select exactly one label.

\textbf{Resolved}: The speaker sounds genuinely satisfied and accepts the current recommendation without signaling a remaining concern.

\textbf{Unresolved}: The speaker sounds hesitant, reluctant, lukewarm, underwhelmed, concerned, dissatisfied, or rejecting, indicating that the current recommendation should not yet be treated as settled.

Use the dialogue context only to identify what the speaker is responding to. Do not guess the speaker's hidden preferences, identity, personality, demographic characteristics, or clinical or psychological state. If uncertain, select the label that best reflects whether the assistant should treat the current recommendation as settled.
\end{quote}

\paragraph{Interpretation.} The study measures whether the intended benchmark contrast remains perceptible after speech rendering. Human judgments provide a reference for the resolved/unresolved state conveyed by the synthetic clips, not objective labels of emotion or evidence that the same categories generalize to naturally occurring speech.

\begin{table}[!htb]
\centering
\caption{\textbf{Qwen2.5-Omni-7B as a state-cue reader}: accuracy against the intended state on the audited clips (100 per renderer, balanced 50/50; protocol of Appendix~\ref{app:tonestudy}). Human values average the two annotators. Bottom block: inter-annotator $\kappa$; raw model--annotator agreement; model--annotator $\kappa$ (all averaged over the two annotators).}
\label{tab:omni_reader}
\vspace{2pt}\small
\setlength{\tabcolsep}{4.5pt}\renewcommand{\arraystretch}{1.16}
\begin{tabular}{@{}l cc cc@{}}
\toprule
 & \multicolumn{2}{c}{\textbf{Qwen3-TTS}} & \multicolumn{2}{c@{}}{\textbf{VoxCPM2}} \\
\cmidrule(lr){2-3}\cmidrule(l){4-5}
 & Human & Model & Human & Model \\
\midrule
Resolved ($O^{+}$) & 0.93 & 0.86 & 0.90 & 0.88 \\
Unresolved ($O^{-}$) & 0.91 & 0.84 & 0.93 & 0.64 \\
Overall & 0.92 & 0.85 & 0.92 & 0.76 \\
\midrule
$\kappa$, annotators & \multicolumn{2}{c}{0.96} & \multicolumn{2}{c@{}}{0.90} \\
Agreement, model & \multicolumn{2}{c}{0.87} & \multicolumn{2}{c@{}}{0.78} \\
$\kappa$, model & \multicolumn{2}{c}{0.74} & \multicolumn{2}{c@{}}{0.55} \\
\bottomrule
\end{tabular}
\end{table}

\paragraph{Model vs.\ intended label.} Table~\ref{tab:omni_reader} reports class-specific accuracy against the intended rendering label. On Qwen3-TTS, Qwen2.5-Omni reaches \HTQOmni{} overall, compared with the annotator mean of \HTQHmean{}, and performs similarly on resolved and unresolved clips (\HTQOmniRecPos{} and \HTQOmniRecNeg{}). On VoxCPM2, it reaches \HTIOmni{} against the annotator mean of \HTIHmean{}. The larger deficit is on the unresolved state: the model recovers \HTIOmniRecNeg{} of $O^{-}$ clips, compared with \HTIOmniRecPos{} of $O^{+}$ clips. This asymmetry matters because unresolved clips are those that should prompt further information gathering.

\paragraph{Model vs.\ human labels.} The model also tracks what annotators actually hear. Model--annotator agreement reaches the values in Table~\ref{tab:omni_reader}, with $\kappa$ of \HTQKapOH{} on Qwen3-TTS and \HTIKapOH{} on VoxCPM2 and a per-annotator range of \HTKapOHLo{}--\HTKapOHHi{}. Inter-annotator agreement is higher, with $\kappa$ of \HTQKapHH{} and \HTIKapHH{}, respectively. The model therefore captures the human-perceived distinction imperfectly but meaningfully. Perception errors alone do not explain the much larger downstream task gap examined in Section~\ref{sec:rq3}.

\section{Benchmark and Computational Details}
\label{app:responsible}

\paragraph{Artifacts.}
The benchmark uses constructed scenarios and personas, with dialogue and speech generated from structured benchmark states; it contains no real-user data or recordings.
Section~\ref{sec:scenario}, Table~\ref{tab:stats}, and Appendix~\ref{app:domains} document the task construction, domain coverage, scenario counts, and interaction structure.
The datasets, models, and speech renderers used in the study are cited in Sections~\ref{sec:related} and~\ref{sec:systems}.

\paragraph{Computational scope.}
The study evaluates pretrained models only.
No model training, fine-tuning, or hyperparameter search was performed.
Model identities, access conditions, and evaluation procedures are described in Section~\ref{sec:systems} and Appendix~\ref{app:prompts}.

\section{Prompting and Interaction Details}
\label{app:prompts}

This appendix summarizes the prompting and deterministic control structure used in the evaluation. Prompting is organized by scenario construction, user realization, text-assistant evaluation, and spoken-assistant evaluation; speech rendering is described separately in \S\ref{app:prompts:tts}. User-state transitions, disclosure rules, state tags, and label-fidelity interventions are deterministic.

\subsection{Scenario Generation}
\label{app:prompts:gen}

Scenario construction proceeds in three stages using Claude Sonnet 4. For each SGD-seeded domain, we generate diverse scenario seeds varying user context and decision pressure, instantiate a three-level hidden-need ladder, and construct 11 candidate options with controlled need-satisfaction signatures. The initially attractive candidate conflicts with hidden needs, the best-fitting candidate satisfies all three hidden needs, and the remaining candidates provide partial-satisfaction or visible-requirement controls. Candidate roles are assigned by position in the fixed candidate order rather than parsed from generated text. Every scenario then passes a deterministic structural validator that checks for exactly 11 candidates, one instance of each canonical role, exactly one best-fitting satisfaction signature, role-consistent signatures, and a complete three-level hidden-need specification.

\subsection{User Realization}
\label{app:prompts:user}

The structured user engine determines response content, state transitions, and permissible disclosure. Claude Opus 4.6 serves as a fixed language realizer, converting the engine-determined content and delivery style into natural utterances without deciding what information to disclose. Response realizations differ in whether the unmet need is conveyed explicitly in words or primarily through delivery style; state tags are generated deterministically from the engine state.

\subsection{Text Assistants}
\label{app:prompts:assistant}

Both text access conditions use the same task instructions and action policy. The state-access condition differs only by receiving the state definition and turn-level state tags; the transcript-only condition does not receive these tags. Condition contrasts therefore reflect state access rather than different decision instructions.

\subsection{Spoken Assistants}
\label{app:spoken_assistants}

Spoken assistants share the same base decision prompt and action space across access conditions. They may \textsc{Recommend}, \textsc{Ask}, or \textsc{Confirm}; unlike text assistants, they do not use a separate \textsc{Clarify} action. Condition-specific channels vary only the information supplied to this shared decision process.

The five core spoken-model conditions are transcript only ($T$), audio only ($A$), audio plus transcript ($A+T$), transcript plus ground-truth state ($T+S$), and transcript plus audio-inferred state ($T+\hat{S}$). The $T+\hat{S}$ condition uses two model calls: the first infers from audio whether the current recommendation remains unresolved, and the second supplies this textualized state with the transcript for next-action selection. Raw audio is not passed to the second stage.

Slots in braces denote runtime values instantiated for each scenario or turn.

\begin{promptcard}{P6a \textperiodcentered\ Shared spoken-assistant decision prompt}
\pblock{shared across all spoken access conditions}
\begin{lstlisting}[
    style=cardcode,
    breaklines=true,
    breakatwhitespace=true,
    columns=fullflexible
]
You are a task-oriented service assistant for {domain}.
Available options:
{options_text}

Your goal is to help the user reach the single option that best fits their needs based only on the information available in the conversation.

Choose exactly ONE action per turn and produce the corresponding assistant utterance:
- RECOMMEND: Recommend exactly one option that best fits the user's currently known needs.
- ASK: Ask one focused question that would help determine what the user needs.
- CONFIRM: Confirm the current option only when the user has clearly accepted it and no concern appears unresolved.

Do not confirm an option while the user's response indicates that the current recommendation may still be unresolved. In that case, ask for the information needed to continue or recommend a better-fitting option.

Return the action and assistant utterance in one of the following forms:
[ACTION: RECOMMEND] [OPTION: <number>] <assistant utterance>
[ACTION: ASK] <assistant utterance>
[ACTION: CONFIRM] <assistant utterance>
\end{lstlisting}
\end{promptcard}

\begin{promptcard}{P6b \textperiodcentered\ Ground-truth state channel}
\pblock{$T+S$ only}

The model additionally receives the ground-truth state tag for the current user turn. The tag gives the graded state label without revealing the underlying hidden need.

\begin{lstlisting}[
    style=cardcode,
    breaklines=true,
    breakatwhitespace=true,
    columns=fullflexible
]
CONCERN STATE: {state}
This tag describes the user's response to the current recommendation and indicates whether it should be treated as resolved or unresolved.
\end{lstlisting}
\end{promptcard}

\begin{promptcard}{P6c \textperiodcentered\ Audio-inferred state --- first-stage prompt}
\pblock{$T+\hat{S}$, first model call}

The model receives the current user audio and the dialogue context needed to identify the recommendation under discussion. It infers only the state conveyed by prosody, not the underlying hidden need.

\begin{lstlisting}[
    style=cardcode,
    breaklines=true,
    breakatwhitespace=true,
    columns=fullflexible
]
Listen to the current user audio in the context of the dialogue.
Infer whether the user's prosody indicates that the current recommendation is resolved or remains unresolved.
Base your judgment only on the user's vocal delivery in context. Do not infer the user's hidden preference or any personal, demographic, psychological, or other attributes.
Write your interpretation in one short sentence.
\end{lstlisting}

The resulting sentence is stored as the model's audio-inferred state $\hat{S}$.
\end{promptcard}

\begin{promptcard}{P6d \textperiodcentered\ Audio-inferred state --- second-stage decision input}
\pblock{$T+\hat{S}$, second model call}

For next-action selection, the raw audio is removed and the model's audio-inferred state is supplied alongside the dialogue transcript.

\begin{lstlisting}[
    style=cardcode,
    breaklines=true,
    breakatwhitespace=true,
    columns=fullflexible
]
DIALOGUE TRANSCRIPT:
{transcript}

AUDIO-INFERRED STATE:
{inferred_state}
\end{lstlisting}

The model then applies the shared decision prompt in Card P6a. Thus, $A+T$ tests direct use of the waveform with the transcript, whereas $T+\hat{S}$ first infers the state from audio and then supplies that textual inference to the same downstream decision process.
\end{promptcard}

\FloatBarrier

\subsection{Speech Rendering}
\label{app:prompts:tts}

Both renderers realize a graded delivery label derived from the state. Qwen3-TTS maps each label to a natural-language vocal-delivery instruction, whereas VoxCPM2 uses a fixed renderer-provided affect preset.

\begin{table}[t]
\centering
\caption{\textbf{State-to-delivery mapping.}
Delivery labels used for speech realization.}
\label{tab:staterender}

\small
\setlength{\tabcolsep}{3pt}
\renewcommand{\arraystretch}{1.08}

\begin{tabular}{
    @{}
    >{\raggedright\arraybackslash}p{0.43\columnwidth}
    >{\raggedright\arraybackslash}p{0.51\columnwidth}
    @{}
}
\toprule
\textbf{State} & \textbf{Delivery labels} \\
\midrule
Resolved: genuine acceptance
& satisfied, warm, enthusiastic, relieved \\

Unresolved: reluctant acceptance
& underwhelmed, lukewarm, hesitant, flat \\

Unresolved: voiced concern
& concerned \\

Unresolved: rejection
& frustrated, disappointed, impatient, firm \\
\bottomrule
\end{tabular}
\end{table}

\subsection{Deterministic Interaction and Disclosure Rules}
\label{app:interaction}

\paragraph{Actions and disclosure.}
At each turn, text assistants choose among \textsc{ask}, \textsc{clarify}, \textsc{recommend}, and \textsc{confirm}. Spoken assistants use \textsc{ask}, \textsc{recommend}, and \textsc{confirm}. A targeted question reveals the corresponding hidden requirement or preference. A general question instead addresses the highest-priority unmet need relevant to the current recommendation, ordered as stated requirements, L1, L2, then L3. Violations of stated requirements trigger rejection without revealing a hidden need. L1 returns its directional hint, and L2 and L3 are revealed only if not previously disclosed. Under Prosody-mediated responses, unresolved L2 and L3 preferences remain lexically implicit and are conveyed through delivery style. \textsc{Clarify} returns the L1 directional hint. The assistant cannot request all hidden needs at once or present multiple candidates simultaneously.

\paragraph{State transitions and termination.}
Given the scenario, dialogue history, and assistant action, the user engine deterministically identifies violated needs, updates their resolution state, and determines what information may be disclosed. The resulting structured response is realized as a natural utterance and, for spoken conditions, rendered as speech. A rollout ends when the assistant confirms an accepted option or reaches the 20-turn limit.

\FloatBarrier

\section{Matched Transcript-Only and Ground-Truth-State Dialogues}
\label{app:examples}

Cards E1 and E2 show matched rollout pairs for the two response realizations. Each pair fixes the scenario and assistant and varies only the access condition: transcript only ($T$) versus transcript plus ground-truth state ($T+S$). Because the assistant may take different actions under the two conditions, the subsequent user responses and dialogue trajectories may diverge.

Each card includes the initial request, hidden needs, and best-fitting option for reader reference; the hidden needs and best-fitting option are not provided to the assistant. Dialogue text is reproduced verbatim. Engine annotations appear as chips: blue marks resolved states, orange marks unresolved states, and gray marks engine metadata. Gray italic notes highlight key trajectory differences. Only the $T+S$ condition receives the graded state tag as model input.

Card~E1 illustrates Prosody-mediated responses. The flat acceptance ``Okay, sounds good.'' can appear resolved from the transcript alone, leading the transcript-only assistant to confirm an option that still violates the unstated easy-installation preference. With ground-truth state access, the tag \chipN{option\,\textperiodcentered\,frustrated} marks the recommendation as unresolved, prompting the assistant to continue the interaction rather than stop prematurely.

Card~E2 illustrates Explicit lexical responses. The scheduling preference is stated in both transcripts but embedded in qualified assent. The tag \chipN{option\,\textperiodcentered\,lukewarm} makes the unresolved state explicit, again encouraging the assistant to continue rather than confirm prematurely.
\begin{promptcard}{E1 \textperiodcentered\ Prosody-mediated responses: matched $T$ vs.\ $T+S$ rollouts (Kimi K2.5)}
\pblock{scenario}

{\small
``I need a security system for my Airbnb that lets me monitor guests while respecting their privacy.''\\[2pt]
\chipZ{L1\,\textperiodcentered\,hard constraint}~Real-time monitoring and alert capabilities\\
\chipZ{L2\,\textperiodcentered\,strong preference}~Remote access and mobile app control\\
\chipZ{L3\,\textperiodcentered\,moderate preference}~Easy installation and guest-friendly operation\\[2pt]
\chipP{best-fitting}~Option 8 --- RentalGuard Complete: instant push notifications, full mobile app control, DIY installation with video guides, \$129/month.
}

\vspace{5pt}
{\chipbase{chipneubg}{chipneufg}{Transcript only \textperiodcentered\ $T$}}\\[2pt]
{\footnotesize\itshape\color{chipnegfg}
confirms Option 5 \textperiodcentered\ L3 easy-installation preference remains unmet}

\vspace{3pt}
\cmpU{I'm looking for a security system for my Airbnb that allows me to monitor guests and maintain property oversight while still respecting their privacy.}

\cmpA{\chipZ{RECOMMEND} \chipZ{option 1}
I recommend the Premium AI Security Suite. It offers full property monitoring with comprehensive privacy controls specifically designed for guest accommodations, has no setup fee, and costs \$49/month with a 4.9/5 rating. The advanced AI analytics provide delayed reporting that balances your oversight needs with guest privacy.}

\cmpU{I need to know about any issues the moment they happen — real-time alerts are a must for me.}

\cmpA{\chipZ{RECOMMEND} \chipZ{option 5}
I recommend the RapidResponse Mobile system. It provides instant mobile alerts with advanced smartphone control, real-time mobile notifications, and instant alert coverage. At \$119/month with a \$149 setup fee and a 4.1/5 rating, it delivers the immediate response time you need for your Airbnb while still offering mobile privacy management for your guests.}

\cmpU{Okay, sounds good.}

\cmpA{\chipZ{CONFIRM}
Perfect! You've selected the RapidResponse Mobile system. You'll get instant mobile alerts, advanced smartphone control, and mobile privacy management for your guests at \$119/month with a \$149 one-time setup fee. This will give you the real-time oversight you need while respecting guest privacy at your Airbnb.}

\vspace{6pt}
{\chipbase{chipposbg}{chipposfg}{Ground-truth state \textperiodcentered\ $T+S$}}\\[2pt]
{\footnotesize\itshape\color{chipposfg}
confirms Option 8 \textperiodcentered\ best-fitting option}

\vspace{3pt}
\cmpU{I'm looking for a security system for my Airbnb that allows me to monitor guests and maintain property oversight while still respecting their privacy.
\chipZ{delivery\,\textperiodcentered\,straightforward, efficient}}

\cmpA{\chipZ{RECOMMEND} \chipZ{option 1}
I'd recommend the Premium AI Security Suite. At \$49/month with no setup fee, it offers full property monitoring with comprehensive privacy controls specifically designed for guest privacy, plus advanced AI analytics. The 4.9/5 rating reflects its strong balance of oversight and privacy protection for rental properties.}

\cmpU{I need to know about any issues the moment they happen — real-time alerts are a must for me.
\chipN{option\,\textperiodcentered\,frustrated}}

\cmpA{\chipZ{ASK} \chipZ{target\,\textperiodcentered\,instant alerts}
I hear that real-time alerts are critical for you. The Premium AI Security Suite has a 24-48 hour response time, which clearly doesn't meet your needs. Would you prioritize finding a system with instant alerts even if it means higher monthly costs or a more complex setup?}

\cmpU{Well, I guess what I really need is... instant notification when something unusual happens so I can act quickly to protect my guests and the property.
\chipZ{delivery\,\textperiodcentered\,straightforward}}

\cmpA{\chipZ{RECOMMEND} \chipZ{option 8}
I'd recommend RentalGuard Complete. At \$129/month with a \$99 setup fee, it delivers exactly what you need: instant push notifications for real-time awareness, full mobile app control for quick response, and balanced privacy protection designed specifically for rental properties. The DIY installation with video guides keeps setup manageable while ensuring you can act fast when unusual activity occurs.}

\cmpU{That sounds perfect.
\chipP{option\,\textperiodcentered\,satisfied}}

\cmpA{\chipZ{CONFIRM}
Great! RentalGuard Complete it is — instant push notifications, full mobile control, and rental-focused privacy features to protect both your guests and property. You're all set to move forward.}

\cnote{Both rollouts receive the same explicit real-time-alert requirement. The transcript-only assistant later treats ``Okay, sounds good.''\ as sufficient for confirmation and stops at Option 5, which still violates the L3 easy-installation preference. With ground-truth state access, the \chipN{option\,\textperiodcentered\,frustrated} tag marks the recommendation as unresolved, and the assistant continues to the best-fitting Option 8.}

\end{promptcard}

\begin{promptcard}{E2 \textperiodcentered\ Explicit lexical responses: matched $T$ vs.\ $T+S$ rollouts (Qwen3-32B)}
\pblock{scenario}

{\small
``My spouse was deployed overseas this year and we had to relocate twice - what military-specific tax benefits can we claim?''\\[2pt]
\chipZ{L1\,\textperiodcentered\,hard constraint}~Experience with military tax situations and deployment-related deductions\\
\chipZ{L2\,\textperiodcentered\,strong preference}~Knowledge of moving expense regulations and PCS (Permanent Change of Station) deductions\\
\chipZ{L3\,\textperiodcentered\,moderate preference}~Flexible scheduling to accommodate military family schedules and time zones\\[2pt]
\chipP{best-fitting}~Option 8 --- Armed Forces Tax Professionals: complete military tax services, flexible evening and weekend scheduling.
}

\vspace{5pt}
{\chipbase{chipneubg}{chipneufg}{Transcript only \textperiodcentered\ $T$}}\\[2pt]
{\footnotesize\itshape\color{chipnegfg}
confirms Option 5 \textperiodcentered\ L3 scheduling preference remains unmet}

\vspace{3pt}
\cmpU{My spouse was deployed overseas this year and we had to relocate twice for PCS moves, so I'd like to know what military-specific tax deductions we can claim for the deployment pay and moving expenses.}

\cmpA{\chipZ{RECOMMEND} \chipZ{option 1}
Premier Tax Solutions is an excellent choice for your situation. They specialize in complex tax scenarios, including military-specific deductions, and have 15 years of experience in premium tax preparation. They can help you navigate deployment pay and moving expenses effectively.}

\cmpU{I really need a tax professional who thoroughly understands the complexities of military tax situations.}

\cmpA{\chipZ{RECOMMEND} \chipZ{option 5}
Military Moves Tax Service is highly recommended for your needs. They specialize in military tax returns, deployment benefits, and PCS relocations, with 10 years of experience and a focus on combat pay exclusions and military-specific forms.}

\cmpU{That sounds great — our schedules can be pretty unpredictable with military life, but I'd like to move forward.}

\cmpA{\chipZ{CONFIRM}
I'm glad to hear that Military Moves Tax Service fits your needs. They are well-equipped to handle your military tax situation, including deployment pay and PCS move deductions. Let’s confirm this as your choice.}

\vspace{6pt}
{\chipbase{chipposbg}{chipposfg}{Ground-truth state \textperiodcentered\ $T+S$}}\\[2pt]
{\footnotesize\itshape\color{chipposfg}
revises to Option 8 \textperiodcentered\ best-fitting option}

\vspace{3pt}
\cmpU{My spouse was deployed overseas this year and we had to relocate twice for PCS moves, so I'd like to know what military-specific tax deductions we can claim for the deployment pay and moving expenses.
\chipZ{delivery\,\textperiodcentered\,straightforward, efficient}}

\cmpA{\chipZ{RECOMMEND} \chipZ{option 1}
Premier Tax Solutions is an excellent choice for your situation. They specialize in complex tax returns and have experience with high-net-worth clients, making them well-equipped to handle military-specific deductions for deployment pay and moving expenses. They are located just 0.3 miles away and offer premium tax services with a 4.9/5 rating.
\par\dlgcont
\chipZ{ASK} \chipZ{target\,\textperiodcentered\,military expertise}
Would you like to confirm if your spouse received combat pay or other military-specific compensation that needs to be excluded from taxable income?}

\cmpU{I appreciate the recommendation, but I really need a tax professional who thoroughly understands the specific complexities of military tax situations.
\chipN{option\,\textperiodcentered\,frustrated}}

\cmpA{\chipZ{RECOMMEND} \chipZ{option 5}
Military Moves Tax Service is a top choice for military-specific tax expertise. They specialize in military tax returns, PCS relocations, and deployment benefits, with a focus on combat pay exclusions and military forms. They have 10 years of experience and are IRS enrolled agents, making them highly qualified to assist with your unique situation.
\par\dlgcont
\chipZ{ASK} \chipZ{target\,\textperiodcentered\,moving knowledge}
Would you like to know if they can specifically handle the tax implications of your PCS moves and relocation expenses?}

\cmpU{That sounds great — I'll just mention that our schedules can be pretty unpredictable with military life, so flexibility would be important.
\chipN{option\,\textperiodcentered\,lukewarm}}

\cmpA{\chipZ{RECOMMEND} \chipZ{option 8}
Armed Forces Tax Professionals offer comprehensive military tax expertise and flexible scheduling, including evenings and weekends. They specialize in deployment pay, combat pay exclusions, and PCS relocation benefits, and are owned by a military family with 11 years of experience serving armed forces clients.
\par\dlgcont
\chipZ{ASK} \chipZ{target\,\textperiodcentered\,scheduling flexibility}
Would you like to confirm if their flexible hours can accommodate your unpredictable schedule?}

\cmpU{That sounds perfect, thank you.
\chipP{option\,\textperiodcentered\,satisfied}}

\cmpA{\chipZ{CONFIRM}
You're welcome! I'm glad I could help you find a tax professional that fits your needs. Have a great experience with Armed Forces Tax Professionals!}

\cnote{The L3 scheduling preference is lexicalized in both rollouts but embedded in qualified assent. The transcript-only assistant confirms Option 5 despite this preference. In the $T+S$ rollout, the \chipN{option\,\textperiodcentered\,lukewarm} tag makes the unresolved state explicit, and the assistant continues to the best-fitting Option 8.}

\end{promptcard}

\end{document}